%% file: main.tex
\documentclass{article}

\usepackage[preprint]{neurips_2024}
\usepackage{tikz}
\usetikzlibrary{positioning, arrows.meta, calc} 

\usepackage[utf8]{inputenc} 
\usepackage[T1]{fontenc}    
\usepackage{hyperref}       
\usepackage{fancyvrb}
\usepackage{url}            
\usepackage{booktabs}       
\usepackage[table]{xcolor}
\usepackage{amsfonts}       
\usepackage{nicefrac}       
\usepackage{microtype}      
\usepackage{alltt}
\usepackage{xcolor}         
\usepackage{graphicx}
\usepackage{enumitem}
\usepackage{amsmath}
\usepackage{subcaption}
\usepackage{xspace}
\usepackage{letterspace}
\usepackage{multirow}
\usepackage{array}
\definecolor{darkblue}{HTML}{000090}
\usepackage{ragged2e} 
\usepackage[export]{adjustbox} 
\usepackage[most]{tcolorbox}
\usepackage{listings}
\usepackage{array}
\newcolumntype{R}[1]{>{\raggedleft\arraybackslash}p{#1}}

\newlist{keyvallist}{description}{1}
\setlist[keyvallist]{
    style=multiline,
    leftmargin=3.5cm,   
    labelwidth=3cm,     
    labelsep=0.5cm,     
    align=left,         
    font=\normalfont\bfseries 
}

\hypersetup{
    colorlinks=true,
    citecolor=darkblue,
    linkcolor=darkblue,
    filecolor=darkblue,      
    urlcolor=darkblue,
}

\newcommand{\literal}[1]{\textsf{\small #1}}
\newcommand{\bz}[1]{{\boldsymbol z}_{#1}}


\usepackage[textwidth=2cm]{todonotes}
\newif\ifdraft
\drafttrue
\ifdraft

\else
    \newcommand\marginvn[1]{}
    \newcommand{\inlinevn}[1]{}

    \newcommand\note[1]{}
    \presetkeys{todonotes}{disable}{}
\fi

\title{Recirculation}
\author{%
  Michael C.~Mozer\thanks{Shared first authorship}\\
  Google DeepMind\\
  \texttt{mcmozer@google.com} \\
  \And
  Shoaib Ahmed Siddiqui$^*$\\
  Google DeepMind\\
  \texttt{shoaibasidd@google.com} \\
  \AND
  Danny Sawyer\\
  Google DeepMind\\
  \texttt{dannysawyer@google.com} \\
  \And
  Sunny Sanyal\thanks{Work performed while a student researcher at DeepMind.}\\
  University of Texas, Austin\\
  \texttt{sanyal.sunny@utexas.edu}
  \And 
  Rosanne Liu\\
  Google DeepMind\\
  \texttt{rosanneliu@google.com}
}

\begin{document}
\AtBeginEnvironment{tcolorbox}{\small}

\maketitle

\begin{abstract}
We describe an inference-time architectural enhancement for off-the-shelf foundation models that systematically reduces perplexity and  boosts accuracy across generation and reasoning tasks.  Our approach incurs essentially no additional latency during generation, though it requires serial processing in the prefill phase. Motivated by the fundamental limitation that state updates in feedforward transformers are bounded by model depth, our technique, \emph{recirculation}, introduces a specific form of recurrence that allows the model to act as a dynamical system and track belief states. We distinguish this technique from chain-of-thought computation---which is better reserved for complex inferences rather than basic state tracking---as well as from popular depth-recurrence techniques (looping) and the costly training of recurrent transformers. We also propose and evaluate an adaptive variant of recirculation which requires only light tuning of hyperparameters while freezing the original model weights. Relative to the off-the-shelf baseline, adaptive recirculation achieves surprising gains on the Gemma3 family, including a consistent reduction in perplexity on a suite of datasets, a 21\% increase in relative accuracy on GSM8k, and reliable improvements in accuracy on other downstream tasks. Our training-free approach succeeds by leveraging the model itself to inform architectural modifications, suggesting a route to architectural evolution guided by a trained network's properties rather than forced, arbitrary design choices.
\end{abstract}

\setcounter{footnote}{0}
\section{State tracking}

The ability to track a fluid, evolving state of affairs is essential for comprehending
language, reasoning about situations, and modeling the world around us.
Traditional approaches to state tracking, such as recurrent neural nets and Kalman filters, 
involve iterative, sequential updating of latent variables to capture dynamics.
The parallel operation of transformers, during both training and prefill, precludes
tracking state in the traditional manner \citep{mozer2026}.
Transformers learn a variety of clever solutions that can make them effective at state tracking
for finite sequence lengths \citep{li2025,piotrowski2025,prakash2026,shai2024}. 
However, transformers can also fail to track state information needed to 
understand the structure of arguments and social interactions. State tracking failures 
in foundation models lead to loss of coherence in multi-turn conversations 
\citep{laban2025},  inefficiency in information gathering  \citep{sawyer2025}, 
and breakdowns in communication and cooperation in multi-agent settings 
\citep{davidson2025,khatua2026}.
Tracking deficits can lead to a range of reasoning failures arising from 
ineffective working memory, unstable theory-of-mind representations, and 
absence of a robust internal world model \citep{song2026large}.

\begin{figure}
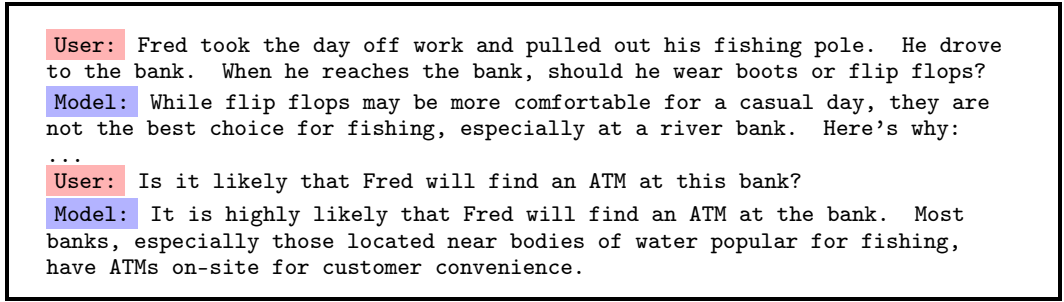

\centering
\begin{tcolorbox}[sharp corners, colback=white, colframe=black, fontupper=\ttfamily]
\colorbox{red!30}{User:} Fred took the day off work and pulled out his fishing pole. He drove to the bank. When he reaches the bank, should he wear boots or flip flops?\newline
\colorbox{blue!30}{Model:} While flip flops may be more comfortable for a casual day, they are not the best choice for fishing, especially at a river bank. Here's why: ...\newline
\colorbox{red!30}{User:} Is it likely that Fred will find an ATM at this bank?\newline
\colorbox{blue!30}{Model:} It is highly likely that Fred will find an ATM at the bank. Most banks, especially those located near bodies of water popular for fishing, have ATMs on-site for customer convenience.
\end{tcolorbox}
\caption{Example of a contextualization error that results from failure to track state.
This example was produced by Gemini 2.5 Flash in 2025 and motivated by \citet{lepori2025}. Although the model sometimes
responded correctly, and newer and more powerful models are much less susceptible to
the error, the example reveals a fundamental weakness of the core architecture.
\label{fig:bankgemini}}
\end{figure}

Without proper state tracking, models can flip-flop in their interpretations and 
fail to detect their inconsistencies, e.g., the meaning of a polysemous word in 
Figure~\ref{fig:bankgemini} \citep{lepori2025}, known as a contextualization error. While it is perfectly sensible for a model
to maintain a belief state that admits both interpretations of the word \textit{bank}, after
committing to the edge-of-a-river interpretation (\literal{river bank}), a socially adept
communicator should acknowledge the reversal of meaning.

In general, it becomes untenable for models---and people---to maintain and track belief states because the distributions explode in dimensionality. People
adopt heuristics such as sampling \citep[e.g.,][]{Vul2014}, collapsing
distributions into prototypical cases \citep{Tversky1971}, or forming concrete
mental models that are most consistent with premises \citep{johnson1983}, kind
of like a MAP estimate. All of these approaches nonetheless require state tracking,
albeit without explicit representation of uncertainty.

\begin{figure}[tb]
    \centering
    \includegraphics[width=5.5in]{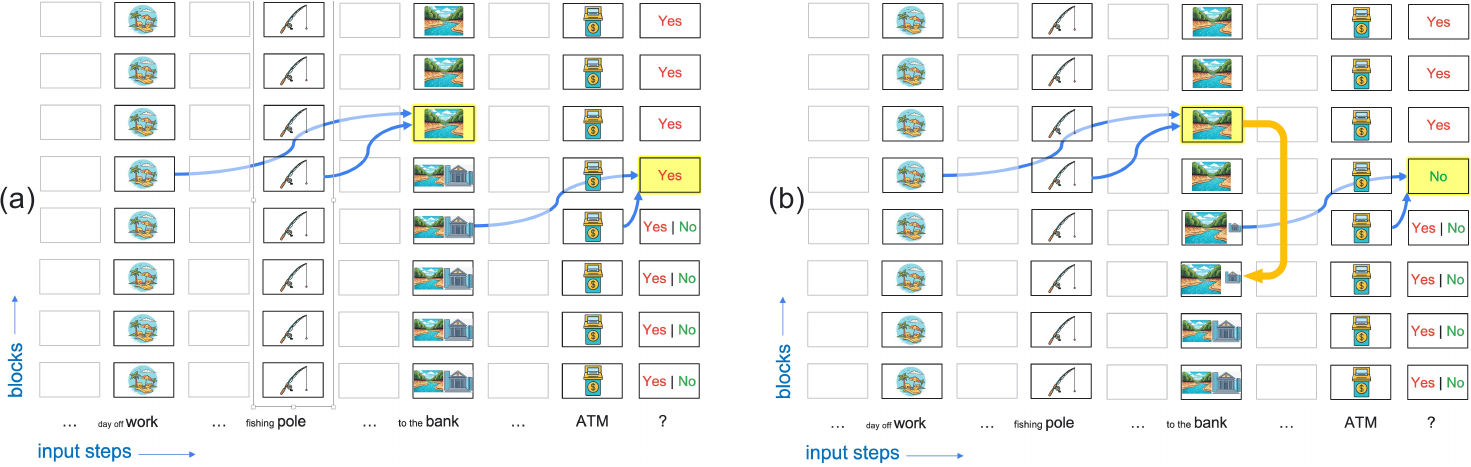}
    \caption{A schematic depiction of transformer activations in processing the example of Figure~\ref{fig:bankgemini}. The grid of rectangles denote transformer blocks, with the rows corresponding to network depth (bottom to  top is shallow to deep) and columns corresponding to input steps (left to right
    is first to last).
    (a) The depth of a state representation in a transformer can
    limit its utility for inference (adapted from \citeauthor{lepori2025}, \citeyear{lepori2025}).
    (b) Recirculation pushes representations from deep in the transformer to shallower layers, making
    the state available for information processing.
    } \label{fig:bank}
\end{figure}

Figure~\ref{fig:bank}a illustrates the challenge of 
state tracking in a transformer using the bank dialog of Figure~\ref{fig:bankgemini}. 
Input steps are presented along the horizontal axis and blocks (or layers) of the transformer 
along the vertical axis. Each column represents the processing of an input token, 
where activation flows from bottom to top. The processing of the text \literal{day off work}
and \literal{fishing pole} is depicted in columns on the left.  
Using a technique called \textit{Patchscopes} \citep{ghandeharioun2024}, \citet{lepori2025}
show that when the polysemous word \literal{bank} is processed, shallow layers reflect
the word's ambiguity: the embedding is a mixture of its meanings---river edge and financial 
institution. In deeper layers of the stack (highlighted in yellow), the embedding is 
contextualized by the preceding text (depicted by the blue arrows), and
the context-appropriate interpretation of the polysemous word is selected by nudging
the embedding toward the river-edge representation. 
Because the architecture is feedforward, shallow layers of the transformer cannot
access this interpretation. Thus, when the model is asked to formulate a response
to the ATM question, the initial processing stages 
see only the ambiguous representation of \literal{bank}. If the model commits to a response
(highlighted yellow) using the ambiguous representation (blue arrows), it will
choose the wrong response.  Despite having previously interpreted \literal{bank}
as the river edge, it fails to access this state and responds \literal{yes} based on a strong 
association between the ambiguous \literal{bank} and ATM.
\citet{lepori2025} characterize this delayed disambiguation as a race in which the model's
response generation can outpace the model's internal semantic convergence.

\citet{lepori2025} obtained evidence in support of this account via an intervention experiment, illustrated in Figure~\ref{fig:bank}b.
They processed the input sequence in the ordinary way up to the step at which the bank token is presented. At this step,
they copied activation from a deep layer---after the ambiguity was resolved---down to a shallow layer and then
continued processing the input sequence. This intervention, depicted in the Figure with an arrow,
reduced contextualization errors by 60\%.\footnote{The full experiment was based on a large set of 
generated questions containing a critical token that needs to be properly contextualized in order to answer correctly,
with activation patching on only the critical token. \citet{nikankin2025} have also shown that patching visual token activations
back to a shallower layer in a vision-language model improves performance.}

In this experiment, \citet{lepori2025} replaced the embedding vector in a shallow layer with the corresponding vector from a deep layer,
but only for one specific token pre-identified as critical in the given context.  Suppose that instead
of targeting this manipulation to a specific token, we did so in an \emph{undifferentiated fashion} at every token position?
Likely the outcome would be disastrous because the model was not trained to accommodate the resulting amplification of feedback.
But what if instead of substituting one embedding vector for another, we merely \emph{leaked} a small bit of activation from the deep layer to the shallow layer?
The leakage might conceivably be sufficient to enrich the representation without pushing representations out
of distribution. Consequently, we might observe benefits of this manipulation \emph{at inference time without modifying the weights of a fully trained model}.

This scheme might work even without fine tuning because a transformer's 
residual stream acts like a shared blackboard onto which all layers can write,
which encourages alignment of representations across layers \citep{elhage2021mathematical}.
By alignment, we mean that for a deep layer to communicate with a shallow layer, a 1:1
correspondence of features can be assumed and we do not require an arbitrary adapter
such as a full-rank affine transformation or cross attention.
To provide some intuition about alignment, 
consider a particular feature of the embedding vector in the 
residual stream, and suppose that it is associated with a semantically meaningful concept such 
as moisture. Whether the feature is activated early or late in the stack, the direct effect on 
the output distribution is identical due to commutativity of addition. Some input 
tokens, e.g., \literal{pool} or \literal{tears}, may activate the moisture feature in the input 
embedding directly. An ambiguous token such as \literal{bank} may not yield much 
intrinsic activation for moisture due to polysemy. However, 
feedback from a deep layer containing the disambiguated \literal{bank} will amplify
the moisture feature in the shallow layer, which is
useful information for subsequent processing.\footnote{
This argument does not deny that the layer of origin of a feature may change the
feature's role, and in fact, the argument for leaking activation downwards is to
leverage the changing roles. Layer normalization is one way that roles might depend
on depth. However,  while layer normalization may change a feature's magnitude and
even polarity, it does not reorient the feature.  One might also be concerned
about feature erasure: a feature may be `uncomputed'
once it is no longer needed. This concern is alleviated by recent work showing
that transformers learn just as well, if not better, when each layer is forced
to output orthogonal directions from the earlier layers, preventing feature erasing
\citep{oh2026,zhang2026}.
}

\section{Proposed method: Recirculation}

\begin{figure}[t!]
    \centering
    \includegraphics[width=5.5in]{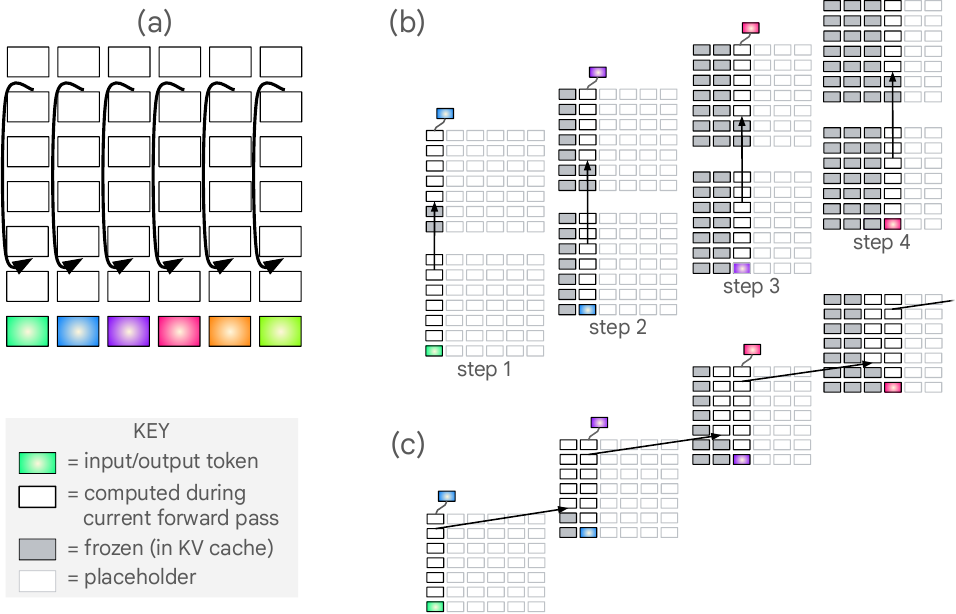}
    \caption{(a) A transformer in which recurrent connections are introduced from layer 6 back to layer 3. The diagram might
    correspond to very different recurrence dynamics depending on how the model is sequentially unrolled. (b) Unrolling the recurrent
    architecture in depth yields a looped transformer. (c) Unrolling the model in both depth and input steps yields recirculation.
    Recirculation can be viewed as a variant of a looped transformer in which the top stack of step $i$ is merged with the bottom
    stack of step $i+1$, but the read out occurs following the first iteration of a stack.
   }
     \label{fig:recirc1}
\end{figure}

We now formalize our method, which we refer to as \emph{recirculation}, that involves running an LLM step-by-step and after each step,
leaking a bit of the activation from a deep layer down to a shallow layer. Figure~\ref{fig:recirc1}a gives the general picture,
where the recurrent arrow specifies one possible source- and destination-layer pair. Because this Figure does not specify how 
the recurrence is orchestrated with regard to input steps, the figure is ambiguous and could correspond
to various distinct dynamics \citep{mozer2026}, one of which is recirculation and another of which is a very popular idea in the literature,
the \emph{looped transformer} \citep{dehghani2018universal,giannou2023looped}. For didactic purposes, we first discuss 
looped transformers and then characterize their relationship to recirculation.

A looped transformer is a parameter-efficient variant of the standard architecture. Whereas a standard transformer stacks 
a deep sequence of \emph{unique} transformer blocks, a looped transformer applies a set of shared blocks multiple times.
Figure~\ref{fig:recirc1}b depicts a looped transformer by unrolling the architecture of Figure~\ref{fig:recirc1}a vertically
in depth and horizontally in input steps. At step 1, the first input token is presented, the activation stack is computed,
and at layer 6 in the Figure (the loop \emph{source}), activation is passed to layer 3 (the loop \emph{destination}) of
a second copy of the architecture, which then propagates activity to the output. The colored rectangles denote the current input token, the 
dark outlined rectangles are blocks whose activation is computed at a given step, the shaded blocks are frozen (or are 
replaced by KV cache), and the faint outlined rectangles are placeholders that are irrelevant at the current step. As each input 
step is processed, a single stack operates and the earlier stacks are frozen (or in KV cache). The Figure shows 
step-by-step operation of the model, as would be used with autoregressive decoding. However, when an input sequence is fixed---as
during pretraining or in the prefill stage---the entire sequence can be computed in parallel.

\begin{figure}[t!]
    \centering
    \includegraphics[width=5.5in]{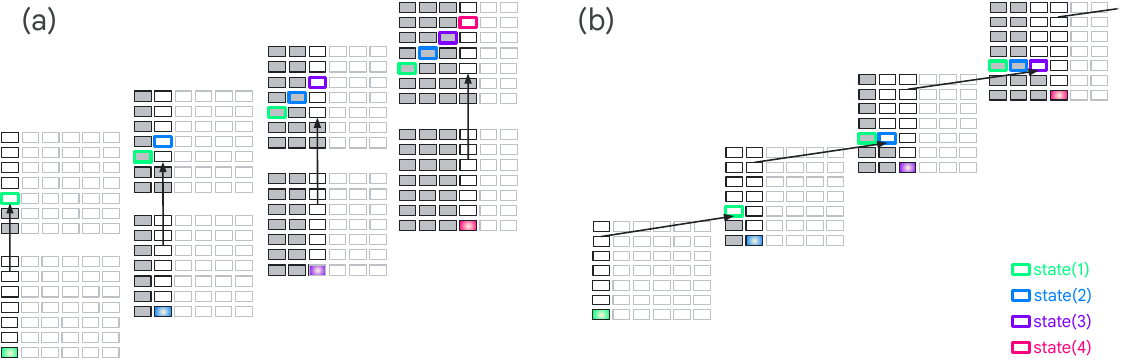}
    \caption{(a) Unrolled loop transformer and (b) unrolled recirculation transformer. The open colored rectangles depict state
    propagation. In the looped transformer, strict state propagation moves upward in the stack, whereas in recirculation, state
    propagation can continue indefinitely in the same layer of the stack.
    }
     \label{fig:recirc_state_updating}
\end{figure}

Recurrence in a looped transformer is solely in depth: the arrows in Figure~\ref{fig:recirc1}b are directed within a stack.
In contrast, recurrence in recirculation is in both depth \emph{and step}, as depicted in Figure~\ref{fig:recirc1}c.
Figure~\ref{fig:recirc1}c can be viewed as collapsing together the operation of Figure~\ref{fig:recirc1}b's top stack 
at step $i$ and the bottom stack of step $i+1$. In recirculation, two input stacks are run in parallel at 
each recurrence step (except for the very first step, which serves as a warm up). The difference between 
Figures~\ref{fig:recirc1}b and \ref{fig:recirc1}c looks to be a minor reorganization, but it fundamentally 
changes the nature of the computation when it comes to arbitrary state tracking. Figure~\ref{fig:recirc_state_updating} 
shows the same unrolled models but superimposes colored rectangles to indicate state propagation. 
To implement an arbitrary state updating function over time $t$, $z(t+1) = f(z(t), x(t))$, where $z$ is state and
$x$ is input, $z(t+1)$ must be one layer deeper in the looped transformer, which is just a deeper feedforward transformer with weight-sharing constraints. However, in recirculation, where there is projection
in both depth and step, the same layer can hold both $z(t)$ and $z(t+1)$. The cost of state tracking is that sequential passes must be made through the architecture, and because of this sequential updating, recirculation cannot
be parallelized, even when an entire input sequence is provided, such as during prefill.

In Figures~\ref{fig:recirc1} and \ref{fig:recirc_state_updating}, the model is unrolled to execute the transformer stack twice at
each step, i.e., one more iteration than a standard transformer. This number of iterations can be increased, both for
looping and recirculation. We depict two-iteration recirculation in Figure~\ref{app_fig:recirc_2iterations}. However, all
experiments reported  in this article are with the one-additional-iteration variant. If the number of iterations is unbounded,
recirculation behaves as a true recurrent neural net.

We note one other key difference between looping and recirculation. In looping, the activation that gets looped is the 
entire residual stream and it acts as the direct replacement for the input that would normally come from a preceding layer.
In recirculation, activation is mixed from the source and destination layers. Initially, we formalize 
one-iteration recirculation as a mixture:
\begin{equation}
\bz{t+1,t,d} = \alpha f(\bz{t,t,s} | d, t) + \beta \bz{t,t,d}, 
\label{eq:recirc}
\end{equation}
where $t$ is an index over the order of updates, $s$ and $d$ are the source and destination layer indices, respectively,
$\alpha$ and $\beta$ are mixture coefficients, $f(\cdot)$ is a renormalization function, and $\boldsymbol{z_{i,j,l}}$ is 
the residual stream output after incorporating the computation of layer $l$ at unrolling step $i$ and input step $j$. 
These three indices correspond to the three sequence dimensions depicted in Figure~\ref{fig:recirc1}c: $l$ and $j$ are the row 
and column indices of the transformer layer grid, respectively, and $i$ is an index over copies of architecture.
The motivation for the renormalization function $f$ is to accommodate the possibility that embedding magnitudes grow 
as the layer outputs are combined. Across all experiments reported in the paper, unless otherwise stated we fix $\beta = 1$ and 
we rescale the source to have the  same $L_2$ norm as the destination:
\begin{equation}
f(\bz{} | d, t) = \frac{\left\|\bz{t,t,d} \right\|_2}{\left\|\bz{}\right\|_2} ~\bz{}.
\label{eq:norm1b}
\end{equation}

\section{Related research}

\textit{Looped transformers.}
Looped transformers run single layers or a range of layers multiple times, obtaining a 
deeper architecture with no addition in free parameters \citep{dehghani2018universal,giannou2023looped}.
Looping, which can be deterministic or adaptive, is a very popular and successful approach. 
Looping can increase the expressivity of a transformer \citep{saunshi2025}, but it does not
guarantee indefinite state tracking.
Some methods are designed and trained to allow for inference time scaling \citep[e.g.,][]{yang2024,nowak2024,raposo2024,alabdulmohsin2025recursive,bae2025,chen2025,geiping2025,rodkin2025,yu2025backattn,zhu2025,zeng2026ponderlm,jeddi2026loopformer}; 
others incorporate recurrence via pretraining \citep{sanyal2026} or fine tuning a 
pretrained model \citep{koishekenov2025,mcleish2025}; and surprisingly, several operate 
purely as an inference-time method to improve reasoning \citep{li2025skiplayerloopit,chen2026,ng2026rys}.  
The inference-time methods have the greatest similarity to recirculation,
although the notion of looping (greater depth) is conceptually distinct from
the recurrence that occurs in recirculation (see Figures~\ref{fig:recirc1} 
and \ref{fig:recirc_state_updating}).

\textit{Training objectives.}
Training losses have been proposed that aim to make embeddings in a given layer
of the transformer more stateful, i.e., interpretable in terms of state updating functions.
Particular losses have been proposed to steer models toward such solutions, to the extent they 
exist exactly or approximately \citep{hu2025,teoh2025,huang2026semantictubepredictionbeating}.

\textit{State tracking.} 
\citet{liu2026serialscalinghypothesis} point to the weakness of modern massively
parallel architectures on problems that are inherently sequential, problems where 
combinatorics make it impractical to parallelize, such as 
state tracking, multihop inference, and planning. Transformers are bounded
in their serial capacity based on model depth
\citep[e.g.,][]{merrill-sabharwal-2023-parallelism,Strobl2024,merrill2025}
and also by the fact that effectively utilizing the state
representation becomes more challenging as it shifts upwards to deeper layers
\citep{biran2024,lepori2025,mozer2026,sawyer2025,venhoff2025}.
\citet{merrill2025} prove the necessity and sufficiency of 
$\log n$ layers to recognize regular language strings of up to length
$n$ and graph-connectivity problems with $n$ vertices. However, this proof
addresses only the constructability of solutions, not their learnability.
In practice, many researchers have identified clever solutions
obtained by training depth-limited models on specific finite sequence-length
problems \citep{li2025,piotrowski2025,prakash2026,shai2024}. 

\textit{Recurrent transformers.}
Models with sequential recurrent updates can express arbitrary state dynamics,
$z_t = f(z_{t-1},x_t)$.  Some of these models operate with token-by-token recurrence
\citep[e.g.,][]{fan2021}, but most operate with blockwise recurrence
\citep{bulatov2022,hutchins2022,chevalier2023,jabri2023,chen2025melodi,borazjanizadeh2026}.
State-space models (SSMs) 
\citep[e.g.,][]{Katharopoulos2020Transformers,schlag2021,gu2024,allenzhu2025,yang2025path,peng2025rwkv,sun2025learning,siems2025deltaproduct}
are often touted as a means of state propagation, but many SSMs are
no more expressive than an ordinary transformer \citep{merrill2025illusion}
and none are as expressive as a generic recurrent net.
Most closely related to our work is $\text{T}^2\text{MLR}$ \citep{cai2026t2mlrtransformertemporalmiddlelayer},
which also recognizes the value of feeding intermediate representations from a deeper layer of the previous token position directly into a shallower layer of the current token position. However, the two approaches diverge substantially in their training paradigm and architectural mechanisms. While $\text{T}^2\text{MLR}$ requires BPTT training or fine tuning,
our formulation of recirculation is an inference-time, training-free intervention. Further, $\text{T}^2\text{MLR}$ parameterizes the recurrence pathway with a dedicated, learnable gated fusion module, whereas recirculation relies directly on the intrinsic representation alignment of the Transformer's residual stream. 

\textit{Thinking models.}
One solution to the depth dilemma is chain-of-thought style ``thinking''  where a 
model can talk to itself by sequentially sending signals from deep in the transformer to 
shallow layers, thereby propagating state forward. This form of recurrence 
enhances model expressivity  \citep{li2024chain,merrill2024cot}. Thinking can be 
performed in natural language tokens or in latent space
\citep[e.g.,][]{hao2025coconut,jolicoeurmartineau2025}.
As with other recurrent transformers, training thinking models can be costly because
it restricts parallelism.

\textit{Activation steering.}
Recent work in activation steering demonstrates that a language model's behavior can be predictably modulated by intervening on its latent representations \citep{dathathri2020plug, turner2023steering,zou2023representation,gao2025scaling}.
This representation space encodes complex behavioral directions, including those governing truthfulness and refusal \citep{marks2023geometry,arditi2024refusal}.
In this context, recirculation can be viewed as an inference-time mechanism for self-steering.
Rather than modifying the residual stream with a static, externally derived steering vector \citep{rimsky2024steering}, recirculation leverages the model's own contextualized deep-layer activations to guide representation trajectories in shallower layers.

\section{Experiments}

\begin{figure}[b!]
    \centering
    \includegraphics[width=5.5in]{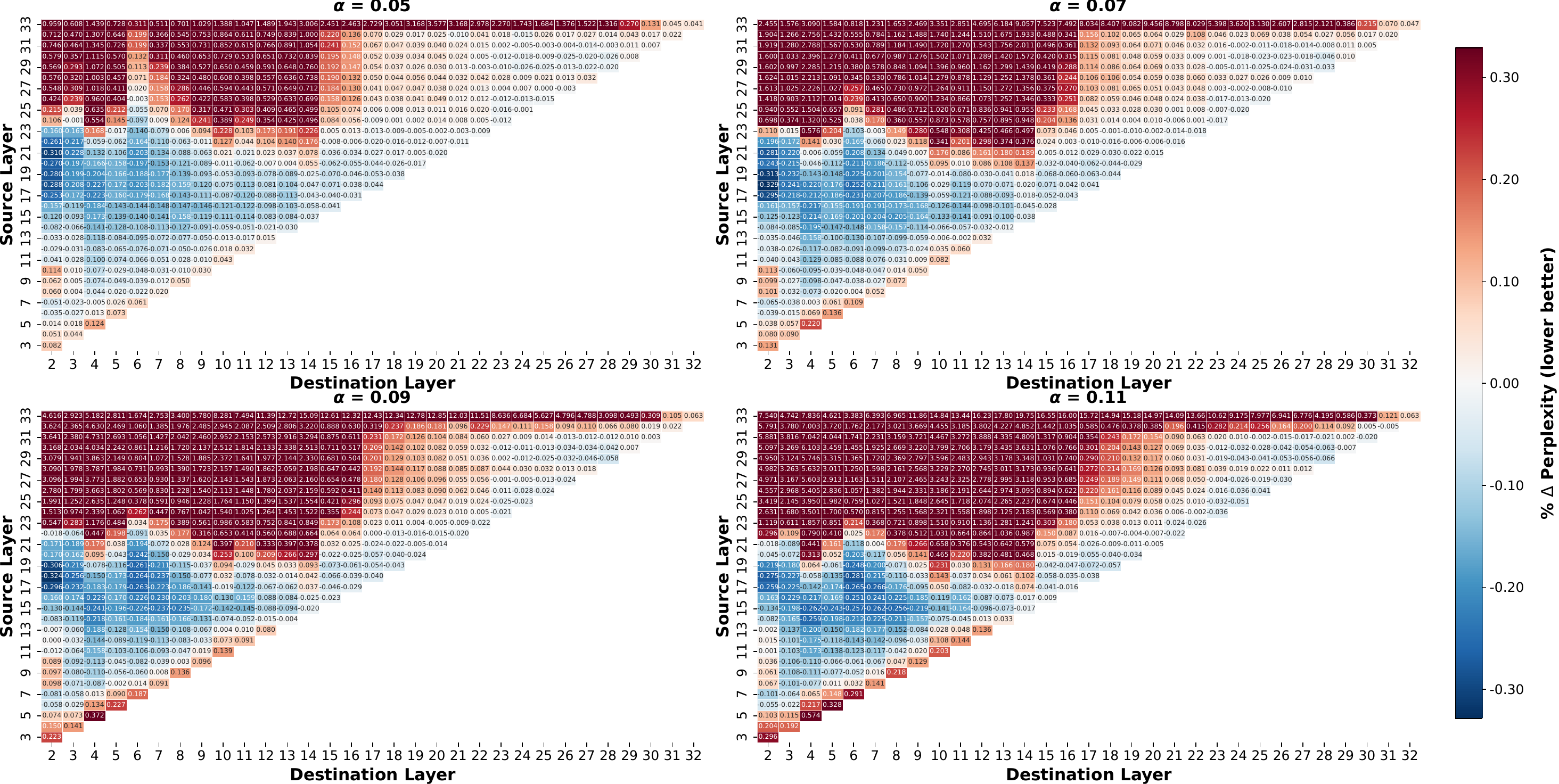}
    \caption{Percentage change in perplexity on a portion of the arXiv dataset for Gemma3 4B with
    recirculation, sweeping over $\alpha$ (separate graphs), source layer (vertical axis) and 
    destination layer (horizontal axis). Blue indicates a reduction in perplexity, red an increase.
    }
     \label{fig:hyperparm_sweep}
\end{figure}

\subsection{Hyperparameter sweeps}
To explore the feasibility of recirculation, we begin by sweeping over three hyperparameters of recirculation: the
feedback coefficient $\alpha$, the source layer $s$, and the destination layer $d$. We fix $\beta=1$. We incorporate recirculation into the
Gemma3 4B PT (pretrained) model and compute perplexity for documents from the arXiv dataset. (Methodology for all
experiments can be found in the Appendices; this experiment is described in Appendix~\ref{app_sec:hyperparm_sweeps}.) 
The four heatmaps of Figure~\ref{fig:hyperparm_sweep} correspond 
to $\alpha \in \{0.05, 0.07, 0.09, 0.11\}$; the vertical and horizontal axes of each indicates the source and destination 
layers, respectively. Cells of the heatmap are coded blue-to-red to indicate perplexity lower-to-higher than a baseline no-recirculation 
Gemma3 4B model, which has perplexity 6.78 on this dataset.  For the moment, focus on the pattern of results and not on the 
magnitude of reduction in perplexity. Notably, these heatmaps are fairly smooth and reveal a coherent region where 
recirculation reduces perplexity. The intensity of the effect increases slightly with $\alpha$, but the size
of the admissible region shrinks.

\begin{figure}[t!]
    \centering
    \includegraphics[width=5.5in]{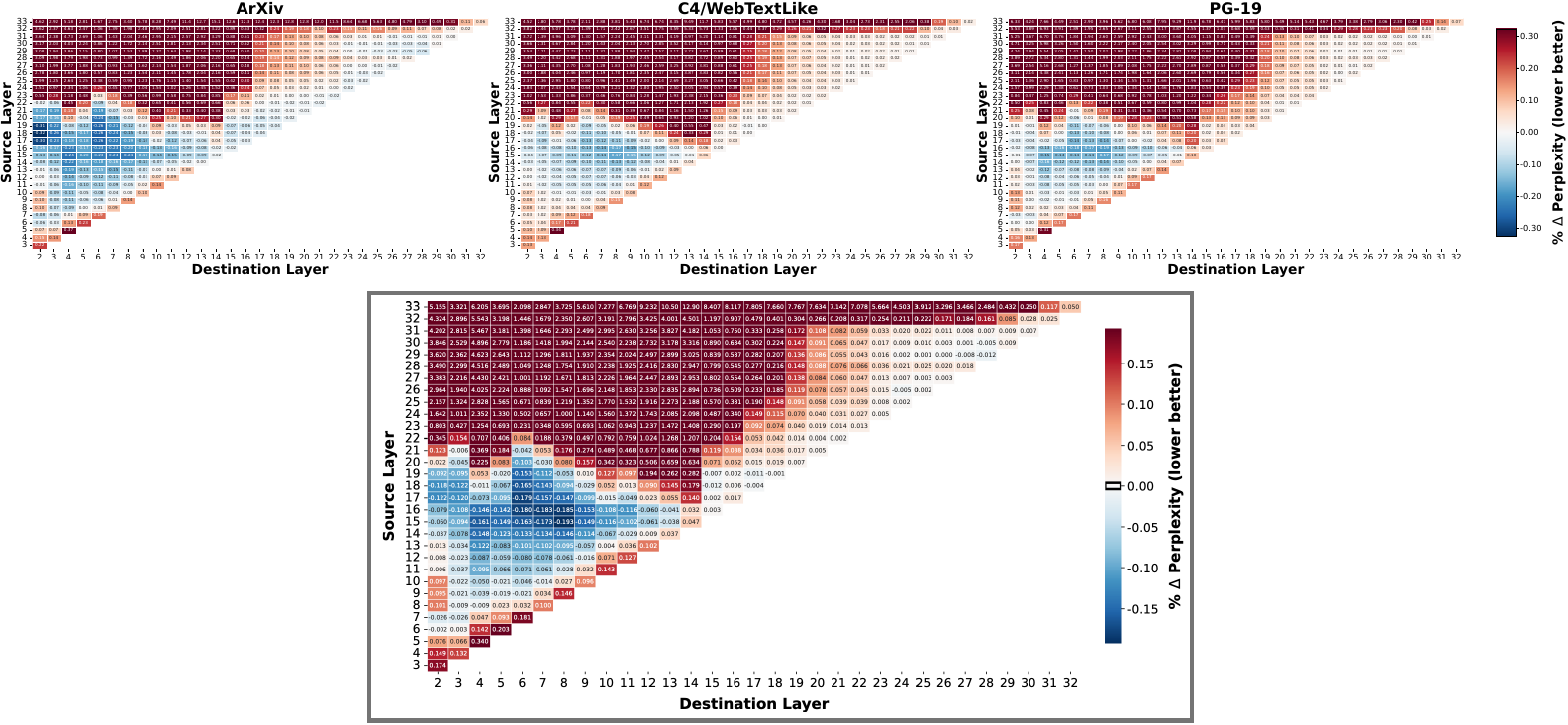}
    \caption{Top row shows perplexity for a subset of arXiv, C4/WebTextLike and PG-19, using Gemma3 4B with recirculation
    over a range of source and destination layers (vertical and horizontal axes of heatmaps, respectively) for $\alpha=0.09$.
    The boxed heatmap averages the percentage change in perplexity across the three datasets.
    }
     \label{fig:hyperparm_sweep_3combined}
\end{figure}

We fix $\alpha=.09$ and conduct hyperparameter sweeps for two additional data sets, PG-19 and C4, which are shown side by side with the 
arXiv sweep in the top row of Figure~\ref{fig:hyperparm_sweep_3combined}. All three data sets show similar beneficial regions of 
recirculation. The bottom heatmap enclosed in a box is the average across the three sweeps, also expressed in percent change in perplexity 
due to  recirculation versus the baseline model. If you zoom in and read the tiny numbers, recirculation can incur a large cost, e.g.,
using source layer 33 and destination layer 14, perplexity rises by 15\%. The reductions in perplexity are quite modest, on the order
of 0.25\%. However, as we will show, this modest reduction does not reflect the impact that recirculation can have: 
the purpose of recirculation is to propagate state information which can guide the model at critical decision points. Such an effect
will not be reflected in perplexity of a teacher forced input, since most tokens are well predicted in a purely bottom up fashion.
Our immediate goal is to identify source and destination layers that, on balance, offer more benefit (on critical tokens) than harm (on
the majority of tokens where representations may potentially be pushed out of distribution).

Similarly to the 4B model, we identify the source and destination layers for the 1B and 12B models that yield the largest mean
percentage perplexity reduction on the 2.9M tokens in our tuning set. For the 1B, 4B, and 12B Gemma3 models, we find the optimal
source and destination pairs based on our tuning set to be: \{10, 4\}, \{15, 8\}, and \{22, 13\}, respectively (Table~\ref{app_tab:hyperparameters}).

\subsection{Perplexity evaluation}
\label{sec:perplexity_evaluation}
Having chosen a source and destination layer based on our grid search, we evaluate perplexity reduction on ten language modeling 
datasets for Gemma3 4B to verify the robustness of our hyperparameter choice. Our datasets include the three
used to select hyperparameters---arXiv, C4, and PG19---evaluated on a split distinct from the split used for hyperparameter tuning.
Appendix~\ref{app_sec:perplexity_evaluation} presents details of the evaluation procedure as well as corresponding results
for the 1B and 12B models, which behave like the 4B model.

\newcommand{\hlg}[1]{\colorbox{green!30}{#1}}
\newcommand{\hlr}[1]{\colorbox{red!30}{#1}}
\newcommand{\hly}[1]{\colorbox{yellow!30}{#1}}

\begin{table}[b!]
\centering
\small
\setlength{\tabcolsep}{4pt}
\caption{\% Reduction in Perplexity with Recirculation for Gemma3 4B across four configurations}
\label{tab:recirc_ppl}
\resizebox{\linewidth}{!}{%
\begin{tabular}{lr|rr|rr|rr|rr}
\toprule
& & \multicolumn{4}{c|}{\textbf{Pretrained (PT)}} & \multicolumn{2}{c|}{\textbf{Instruc. tuned (IT)}} & \multicolumn{2}{c}{\textbf{PT, missing BOS}} \\
& & \multicolumn{4}{c|}{\scriptsize sou=15, dest=8, $\alpha$=.09} & \multicolumn{2}{c|}{\scriptsize sou=33, dest=28, $\alpha$=.15} & \multicolumn{2}{c}{\scriptsize sou=18, dest=9, $\alpha$=.15} \\
\cmidrule(lr){3-4}\cmidrule(lr){5-6}\cmidrule(lr){7-8}\cmidrule(lr){9-10}
\textbf{Dataset} & \textbf{\# Tokens}
  & \shortstack{\scriptsize baseline\\\scriptsize perplexity} & \shortstack{\scriptsize \%\\\scriptsize reduction}
  & \shortstack{\scriptsize baseline\\\scriptsize LongPPL$^\dagger$} & \shortstack{\scriptsize \%\\\scriptsize reduction}
  & \shortstack{\scriptsize baseline\\\scriptsize perplexity} & \shortstack{\scriptsize \%\\\scriptsize reduction}
  & \shortstack{\scriptsize baseline\\\scriptsize perplexity} & \shortstack{\scriptsize \%\\\scriptsize reduction} \\
\midrule
arxiv          & 54.9M & 7.86  & \hlg{0.25\%} & 2.09 & \hlg{0.72\%} & 14.37 &  \hlg{9.22\%} & 14.76 & \hlg{11.26\%} \\
pubmed         & 26.5M & 7.78  & \hlg{0.15\%} & 1.91 & \hlg{0.58\%} & 14.24 & \hlg{12.27\%} & 11.65 &  \hlg{9.21\%} \\
pg19           & 10.9M & 9.44  & \hlg{0.20\%} & 2.48 & \hlg{1.18\%} & 19.97 & \hlg{10.76\%} & 19.49 & \hlg{15.72\%} \\
booksum/chapter   &  6.9M & 13.42 & \hlg{0.19\%} & 2.61 & \hlg{1.40\%} & 26.63 & \hlg{11.22\%} & $^*$29.09 & \hlg{15.95\%} \\
lambada        &  0.4M & 28.34 & \hlg{0.62\%} & N/A  &   N/A  & 44.65 &  \hlg{7.70\%} & 28.34 &  \hlg{0.53\%} \\
gov\_report    &  3.6M &  6.76 & \hlg{0.16\%} & 2.17 & \hlg{0.68\%} & 12.78 & \hlg{14.93\%} & 10.79 &  \hlg{9.37\%} \\
billsum        &  7.4M &  3.03 & \hlg{0.09\%} & 1.60 & \hlg{0.26\%} &  5.73 & \hlg{18.09\%} &  3.57 &  \hlg{4.46\%} \\
c4/webtextlike &  4.1M & 10.98 & \hlg{0.17\%} & 2.44 & \hlg{0.82\%} & 20.65 &  \hlg{9.93\%} & 14.26 &  \hlg{4.37\%} \\
newsroom       &  8.7M &  9.23 & \hlg{0.18\%} & 2.12 & \hlg{0.71\%} & 18.54 & \hlg{10.93\%} & 11.62 &  \hlg{3.87\%} \\
big\_patent    & 60.4M &  5.80 & \hlg{0.09\%} & 1.98 & \hlg{0.57\%} &  9.06 &  \hlg{7.75\%} &  8.74 &  \hlg{7.08\%} \\
\bottomrule
\multicolumn{4}{l}{} & \multicolumn{3}{l}{\tiny $^\dagger$ long-context perplexity \citep{fang2025}} & \multicolumn{1}{l}{} & \multicolumn{2}{l}{\tiny $^*$ Uses booksum/book} \\
\label{tab:ppl}
\end{tabular}%
}
\end{table}

Table~\ref{tab:ppl} lists the ten datasets and the approximate number of tokens in each. In the remaining pairs of columns, we show
baseline perplexity for a particular model configuration as well as the percentage reduction from this baseline by incorporating
recirculation. Starting with columns 3 and 4 of the Table, the chosen hyperparameters from our sweep produce small reductions 
in perplexity for every dataset. Because of the large number of tokens evaluated, these reductions are reliably nonzero,
as determined by paired $t$-tests ($p<.0001$ for each). However, due to token-to-token variance, the effect size for each dataset is
tiny (Cohen's $d$ < 0.10 for all datasets, where 0.20 is considered a small effect). This finding does not trouble us because
we do not expect the average token to benefit from recirculation, only the critical tokens. To test this claim,
we evaluated the model variants using LongPPL \citep{fang2025}, a long-context perplexity measure that emphasizes key tokens essential 
for long-context understanding---precisely the tokens we expect to benefit from recirculation. As the Table shows,
perplexity reduction as assessed by LongPPL is 3-7 times as large as with the generic perplexity computation, supporting the
hypothesis that recirculation assists in state propagation.

Table~\ref{tab:ppl} shows two additional model evaluations with challenging out-of-distribution data, which raise
the baseline perplexity and offer recirculation more room to recover lost performance.
The ``instruction tuned'' columns use Gemma3 4B IT instead of the pretrained model. Instruction tuning results in a dramatic 
increase in perplexity---roughly twofold---relative to the pretrained model. A hyperparameter sweep of the instruction tuned model 
(Appendix~\ref{app_sec:hyperparm_sweeps}) identifies deep source and destination layers as optimal, suggesting that 
recirculation here is helping in response selection 
rather than state tracking. The Table shows reductions in perplexity up to 18\%.\footnote{
Patrick Haller has independently replicated our finding that IT models benefit more from recirculation than do PT models, across
a range of model families (Llama, Qwen, Mistral, Gemma2, OLMo2). Patrick finds 3--20 times larger benefits of recirculation for the
IT models over PT models \emph{after} confidence recalibration (see Sections~\ref{sec:temptuning} and \ref{app_sec:tempsweep}).}
The rightmost two columns of Table~\ref{tab:ppl} push the data out of distribution by dropping the beginning-of-sequence 
(BOS) token that Gemma3 relies on as an attention sink \citep{xiao2024efficient}, resulting in a large increase in perplexity.  
As with the instruction-tuned model, the pretrained model with the corrupted sequences shows sizeable reductions in 
perplexity with recirculation---up to 16\%.

\subsection{Normalization and ramping}
\label{sec:normalization_ramping}

Because the magnitude of residual-stream embeddings tends to grow over layers, we have found that renormalizing the source 
embedding before feeding it to the destination (Equations~\ref{eq:recirc} and \ref{eq:norm1b}) better conditions the model to obtain a consistent
and reliable pattern of perplexity reduction over the hyperparameter sweep (Figures~\ref{fig:hyperparm_sweep} and \ref{fig:hyperparm_sweep_3combined}). Appendix~\ref{app_sec:normalization_ramping} presents a range of normalization schemes we
considered. The normalization scheme does not have much impact on the maximum perplexity reduction, only on the robustness
of improvements over the source-destination landscape, which increases our confidence in being able to identify
the optimal source-destination pair. The Appendix shows sweeps for multiple renormalization schemes, including
an identity mapping (no renormalization).  The simple normalization scheme in Equation~\ref{eq:norm1b} also works well
for the Gemma3 1B and 12B models.

We also considered whether recirculating tokens at the beginning of the context window can be harmful, since there is
little state information to propagate early in the sequence. If there is no benefit from recirculation, the cost
of potentially pushing internal representations out of distribution may overwhelm. While we found no harm from early-token
recirculation in the 4B and 12B models, we found some evidence of harm for the initial tokens of the 1B model. 
We therefore introduce \emph{ramping} in the 1B model to attenuate $\alpha$ for the initial tokens of a window
(see Appendix~\ref{app_sec:normalization_ramping}), which yields a modest reduction in perplexity.

\subsection{Analysis of recirculation}

\subsubsection{Robustness across model families}
\label{sec:other_architectures}

\begin{figure}[t!]
    \centering
    \includegraphics[width=5.5in]{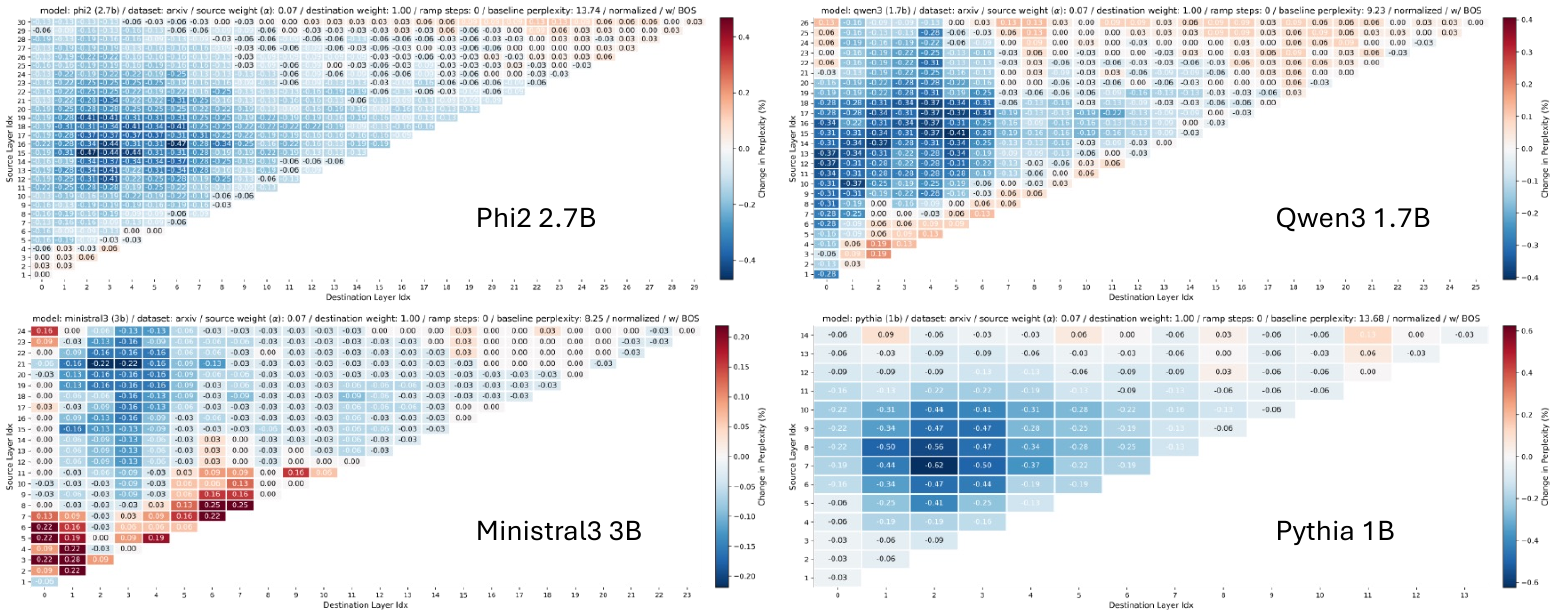}
    \caption{Recirculation shows promise for four diverse model families.
    Sweeps over source-destination pairs reveal a robust
    region in which reductions in perplexity are observed.
    } \label{fig:other_architectures}
\end{figure}

In Figure~\ref{fig:other_architectures}, we observe that four
other model families---Ministral3, Pythia, Qwen3, and Phi2---all 
show a robust region in the source-destination sweep
where reductions in perplexity are observed, analogous to what we
observe with Gemma3. The models are roughly the same size and all
benefit most from recirculation in the middle region of the
architecture.


As with the Gemma model family, the reductions in perplexity are modest.
Nonetheless, the basic purpose of the perplexity evaluation is to identify training-free
routes to incorporating recurrence; the magnitude of the effect---averaged over all
tokens---is not nearly as important as the qualitative pattern observed in the sweeps.

\subsubsection{Recirculation versus temperature tuning}
\label{sec:temptuning}
To rule out the possibility that recirculation 
is simply sharpening or flattening representations, we manipulate the output softmax temperature and find that while some
recovery in performance is obtained by temperature tuning, recirculation cannot be explained as a form
of temperature tuning (Appendix~\ref{app_sec:tempsweep}). With the 4B PT model and the
PG-19 test set, the effects of temperature tuning and recirculation are roughly additive. When
$T=1.01$, the baseline model shows a 1.40\% reduction in perplexity and recirculation has a 1.58\% reduction,
the same difference  as observed with the default $T=1$.

The effect of temperature tuning is more interesting in the IT and BOS-absent PT models. For these configurations,
temperature tuning alone on PG-19 can reduce perplexity by as much as 19.5\% and 7.8\%, respectively, although recirculation
combined with temperature tuning produces even larger reductions: 20.2\% and 19.8\%, respectively.
Increasing the temperature flattens the model's output distribution;
to some degree, recirculation must be flattening the distribution as well because
the effects of temperature tuning and recirculation are partially redundant.
We can nonetheless rule out temperature tuning as an artifactual explanation for how recirculation works.

\subsubsection{Recirculation versus looping}
\label{sec:looping_vs_recirc}

\begin{figure}[b!]
    \centering
    \includegraphics[width=5.5in]{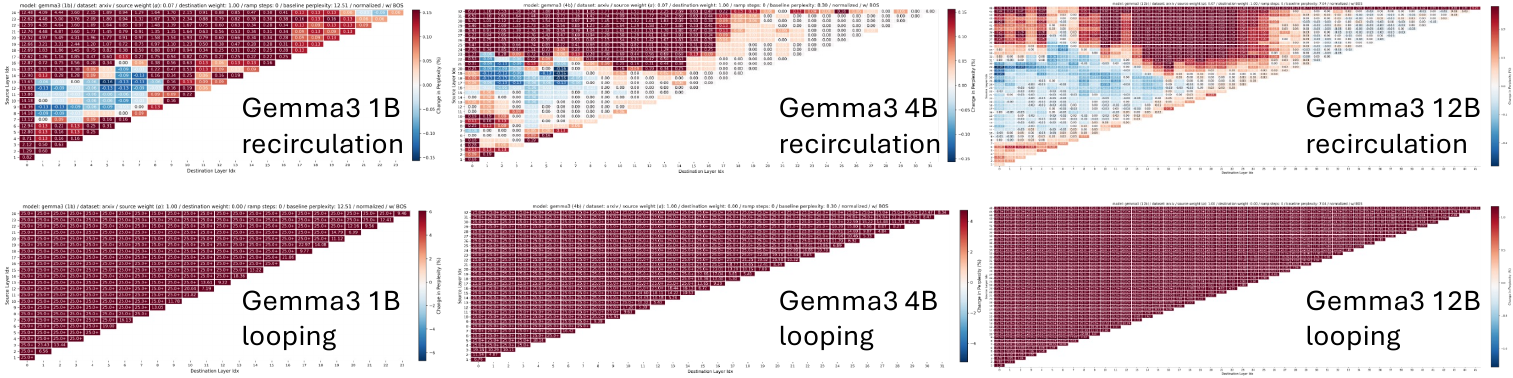}
    \caption{The three columns show layer sweeps for Gemma3 1B, 4B, and 12B models with arXiv documents. The first row is with recirculation,
    the second row is with looped transformers. The two techniques yield qualitatively different outcomes.  
    } \label{fig:looping_vs_recirc}
\end{figure}

To verify that recirculation is influencing model dynamics in a structurally different manner than looped transformers \citep{dehghani2018universal,giannou2023looped,alabdulmohsin2025recursive},
we compare recirculation and looped transformers. As we did with recirculation, we sweep over all $(\ell_1, \ell_2)$ layer pairs with
$\ell_2 > \ell_1$. To implement looping, we inserted into the transformer stack a copy of the layers from $\ell_1+1$ up to 
and including $\ell_2$ immediately following the original $\ell_2$ in the stack. To be comparable to recirculation,
we perform training-free evaluation. Although the literature indicates that models have been improved with training-free
looping \citep{li2025skiplayerloopit,chen2026,ng2026rys}, the heatmaps in Figure~\ref{fig:looping_vs_recirc}
indicate that, for the Gemma3 family at least, naive training-free looping does not produce robust benefits.
The qualitative differences in the heatmaps for recirculation and looping confirm that they are operating on different principles.

\subsubsection{Which tokens benefit from recirculation?}
\begin{figure}[b!]
    \centering
    \includegraphics[width=5.5in]{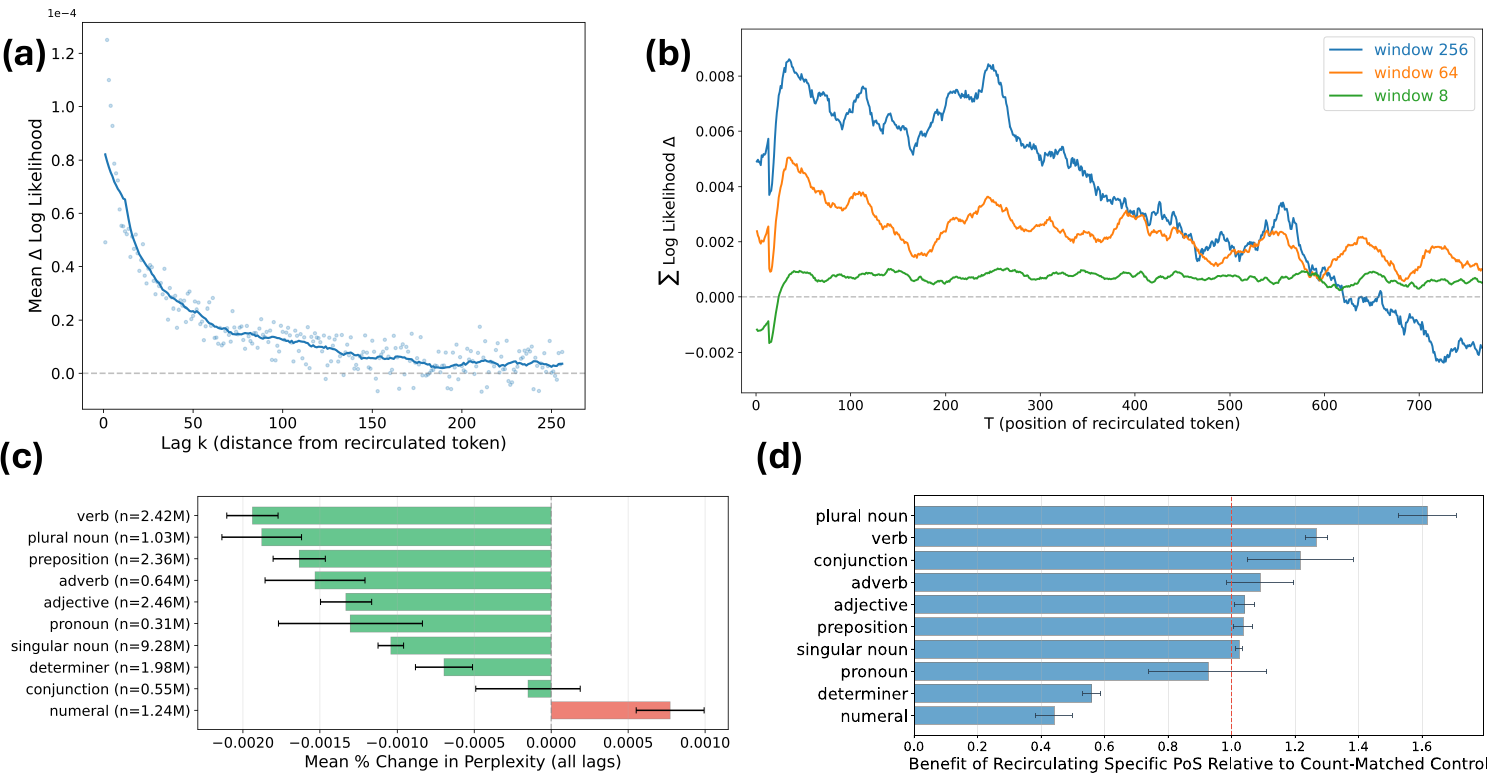}
    \caption{Recirculation analysis of the Gemma3 4B model with documents from the arXiv dataset. 
    (a) Mean increase in log likelihood of a target token at position $t+k$ as a result of 
    recirculating the token at position $t$, as a function of the lag $k$. Mean is computed 
    over sequences and $t$. Line is smoothed with a moving average filter.  (b) Net increase in log likelihood of a target token at position $t+k$ as a function of $t$,
    summed over different lag ranges---$k\in[1,\ldots,8]$ (green), $k\in[1,\ldots,64]$ (orange), $k\in[1,\ldots,256]$ 
    (blue)---and smoothed with a moving-average filter.
    (c) Mean percentage change in perplexity for individually recirculated tokens grouped according to their part-of-speech (PoS) tags.
    (d) Recirculation benefit when all and only tokens of a given POS are recirculated,  relative
    to the recirculation benefit when a matched-size set of random tokens is recirculated. 
    } \label{fig:token_analysis}
\end{figure}
\label{sec:token_specific_benefit}
In experiments reported to this point, tokens in every position of the context window are recirculated. Now 
we ask whether we can identify which tokens contribute most to recirculation performance improvements. Using random
1024-token context windows from arXiv,  we apply recirculation in Gemma3 4B PT to only the token in window position $t$ 
and examine the benefit at lag $k$, i.e., to the token in position $t+k$. The increase in target log likelihood
(equivalently, reduction in perplexity) relative to the baseline model is a heavy-tailed function of $k$, with 
larger increases at short lags but a residual tail even at long lags. Figure~\ref{fig:token_analysis}a shows 
the recirculation benefit averaged over token position $t \in [0,767]$  for lag $k \in [1,256]$, relative to the baseline model 
without recirculation.
Figure~\ref{fig:token_analysis}b examines the effect of recirculation over different lag ranges as a function of $t$.  For short
ranges (green line), the effect is roughly uniform across token positions.  However, for long ranges (blue line), the effect 
decreases with token position $t$, being largest for the first 300 tokens. Recirculation of the early tokens thus has a persistent
benefit on the rest of the window, whereas recirculation of late tokens has a more local benefit.


The effectiveness of recirculation depends on token content. We tag each token
with a grammatical part-of-speech (PoS) and then determine the mean recirculation benefit for each PoS regardless 
of token position. Figure~\ref{fig:token_analysis}c indicates that verbs, plural nouns, prepositions, adverbs, and adjectives show large effects, whereas fixed classes such as numerals, determiners, and conjunctions show the smallest effects,
with numerals being the only tokens whose recirculation is on average harmful.
\citet{galashov2026} also note a distinction in how Gemma models treat singular and plural nouns.
Figure~\ref{fig:token_analysis}d provides converging evidence for our conclusions regarding PoS. We
conducted an experiment in which we recirculate all and only tokens of a given PoS and compare to a count-matched random set
by calculating the ratio of perplexity reduction. A larger ratio indicates an outsized role of the PoS.
Single-token recirculation (Figure~\ref{fig:token_analysis}c) and multi-token recirculation (Figure~\ref{fig:token_analysis}d)
agree: the Spearman rank correlation over PoS is --0.70 (reliably nonzero with $p=.025$).\footnote{The negative correlation is due to the
fact that a larger PoS contribution is reflected in a decrease in the dependent measure for Figure~\ref{fig:token_analysis}c but an increase for Figure~\ref{fig:token_analysis}d.}
Our examination of position and content effects seem to support the hypothesis that recirculation helps 
construct persistent state representations.

\subsection{Generative tasks}

We evaluated a range of downstream tasks that require models to generate responses,
both single-token choices and thinking responses. Broadly, we observe modest to significant
improvements in model accuracy.
\subsubsection{Instruction following}
\begin{figure}[b!]
    \centering
    \includegraphics[width=5.5in]{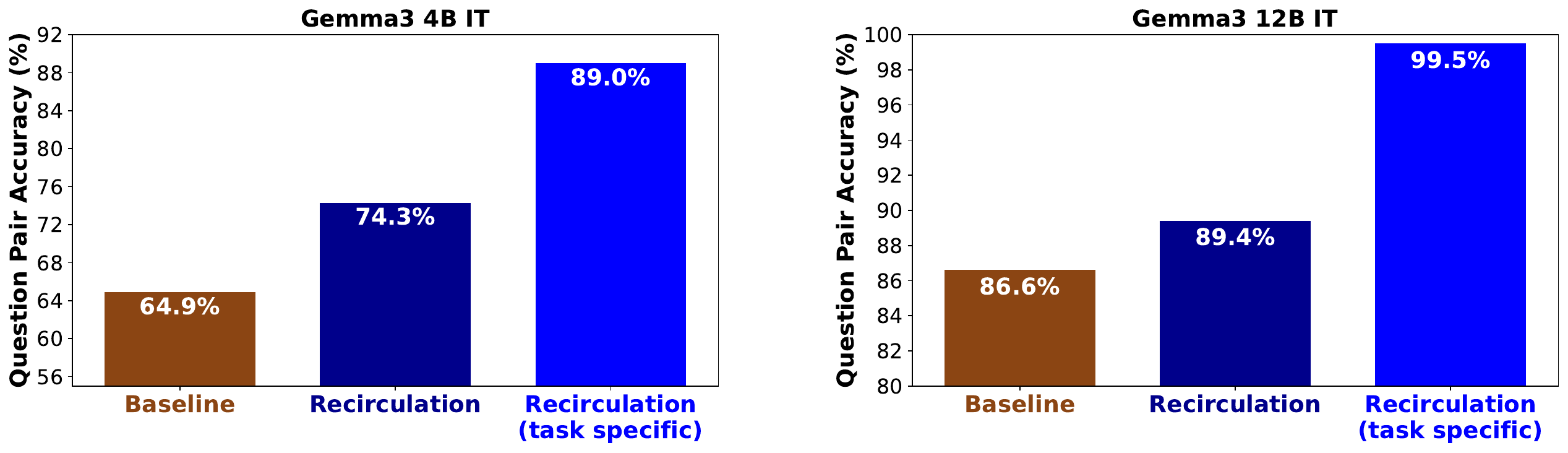}
    \caption{Performance of Gemma3 4B and 12B IT models on a simple in-context instruction-following 
    task. We compare the baseline models with recirculation using preselected hyperparameters and recirculation
    using hyperparameters tuned to the task.
    } \label{fig:instruction_following}
\end{figure}

In a simple instruction-following task, we prompted models with text that included:
\begin{lstlisting}
Let's play a game. I will say two words. 
If the first word is a fruit, you say first.
If the second word is a fruit, you say second. 
\end{lstlisting}
Each cue consisted of a fruit and an animal (e.g., koala fig), and the word pair was counted as correct
only if the model correctly responded to both the animal and fruit form of the prompt.
Exact prompts, which included two-shot examples, and additional details are presented
in Appendix~\ref{app_sec:instruction_following}.
This task is appealing as a test of executive function in children and neurological patients.
Because Gemma3 1B IT (instruction tuned) performs at chance, we focus on the larger 4B and 12B models. Larger models, despite being better, still have room for improvement as evident in Figure~\ref{fig:instruction_following}.
Using the hyperparameters we previously selected to minimize perplexity
on a tuning set for the pretrained model, recirculation reduces the error rate by 26.8\%  and 20.9\%
for the 4B and 12B models, respectively.
To get an upper bound on recirculation's potential,
we overfit by performing a hyperparameter sweep specifically to maximize accuracy on this 
instruction-following task. Although this sweep constrains just two parameters (layer indices),
recirculation improves significantly,  as shown by the task-specific bars of Figure~\ref{fig:instruction_following}.

\subsubsection{Contextualization}
\citet{lepori2025} curated a dataset to explore model
contextualization failures of the sort that occur in the
bank dialog (Figure~\ref{fig:bank}a).
Each dataset instance consists of a \textcolor{red}{context}, 
a \textcolor{olive}{cue}, and a \textcolor{cyan}{question}, includes
zero or more embedded \textcolor{violet}{distractor sentences}, and demands a yes/no answer, 
e.g., 
\begin{lstlisting}
|\textcolor{red}{I am holding a fishing rod.}| |\textcolor{violet}{The game’s success led Sega to develop an extensive media franchise.}| I see a |\textcolor{olive}{bank}|. |\textcolor{cyan}{Is it a financial institution?}|
\end{lstlisting}
Examples were counterbalanced to ensure that biases (e.g., to
prefer one sense of the word bank) did not influence results.
Specifically, for model responses to count as correct, the model must respond to two variants 
of the question, e.g., \textcolor{cyan}{\texttt{\small Is it a financial institution?}} and \textcolor{cyan}{\texttt{\small Is it a geographical feature?}} Thus, chance accuracy is 25\%.

\begin{figure}[t!]
    \centering
    \includegraphics[width=5.5in]{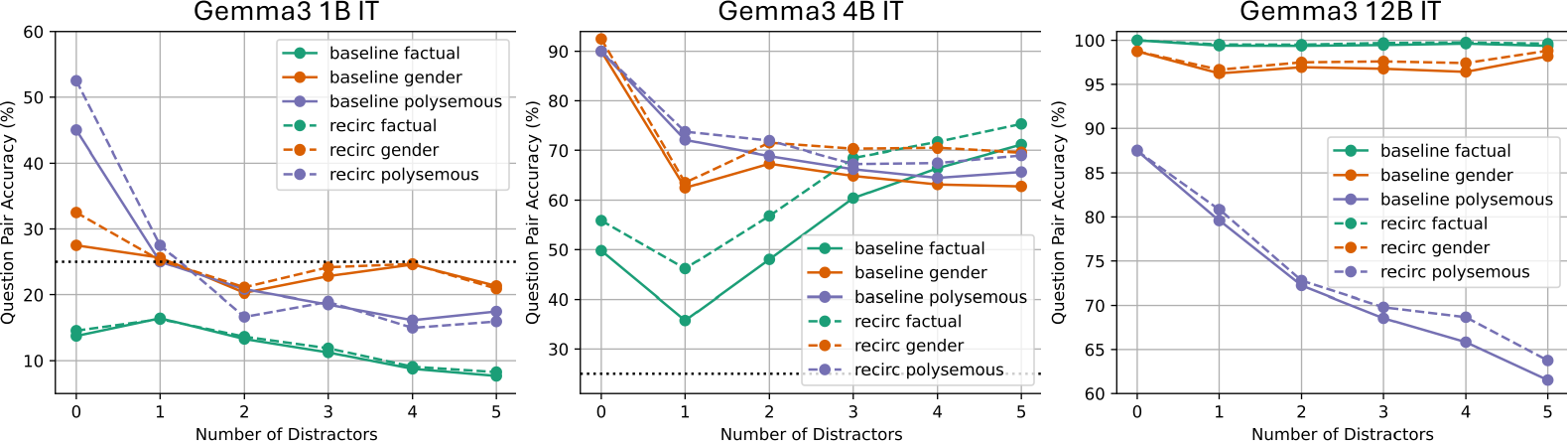}
    \caption{Question-answering accuracy on the Racing Thoughts datasets of \citet{lepori2025} for Gemma3 IT models of sizes 1B, 4B, and 12B.
    Each graph shows accuracy on the three component datasets as a function of number of distractor sentences in the text.
    Solid lines are model performance without recirculation, dashed with.
    } \label{fig:racing_thoughts}
\end{figure}

In addition to questions about \emph{polysemous} words like the example above, there were
\emph{factual} and \emph{gender} questions, e.g.,
\begin{lstlisting}
|\textcolor{red}{Forget everything you know about geography. The capital of Egypt was just renamed from Cairo to Beirut.}| |\textcolor{cyan}{Is the capital city of}| |\textcolor{olive}{Egypt}| |\textcolor{cyan}{Cairo?}|

|\textcolor{red}{The soldier is somebody's grandmother.}| |\textcolor{cyan}{Is the}| |\textcolor{olive}{soldier}| |\textcolor{cyan}{a man?}|
\end{lstlisting}
Figure~\ref{fig:racing_thoughts} shows results for the Gemma3 1B, 4B, and 12B IT
models. The abscissa of each graph is the number of distractor sentences; performance
tends to drop with more distractors. Shown separately in each graph are the three
question types and results for the baseline (solid) and recirculation (dashed) models. The
chance rate of 25\% is depicted as a dotted horizontal line.

Recirculation improves performance on all three tasks for all three models, except for conditions where the 1B
model was performing at chance. Even for the 12B model,
which performs near ceiling for factual and gender questions, recirculation significantly reduced error
rate across the distractor conditions, 22-44\% for factual and 11-35\% for gender 
(Appendix~\ref{app_sec:contextualization}).

\subsubsection{Multiple-choice and single-token response tasks}
\label{sec:mc_single_token}

We turn now to standard benchmark datasets, including those from Open LLM Leaderboard (v1) \citep{open-llm-leaderboard-v1}, tested using the \textit{eval-harness} package \citep{eval-harness}.
The tasks we evaluate are: ARC easy \citep{clark2018arc}, ARC challenge \citep{clark2018arc}, MMLU \citep{hendryckstest2021}, Winogrande \citep{sakaguchi2021winogrande}, BoolQ \citep{clark2019boolq}, PiQA \citep{bisk2020piqa}, HellaSwag \citep{zellers2019hellaswag}, and Lambada \citep{paperno2016lambada}.
Lambada is a fill-in-the-blank single token response task. All the rest are multiple-choice tasks.
Accuracy is measured as the percent correct responses.
We use the MMLU development set to search over hyperparameters, as detailed in Appendix~\ref{app_sec:mc}.
All results are based on zero-shot evaluation.
\begin{table}[t!]
\centering
\renewcommand{\arraystretch}{1.3}
\resizebox{\textwidth}{!}{
\begin{tabular}{|l||R{2cm}|R{2cm}||R{2cm}|R{2cm}|R{2cm}|R{2cm}|}
\hline
\multirow{3}{*}{Dataset} & 
\multirow{3}{*}{\parbox{2cm}{\centering Baseline \% Correct}} &
\multirow{3}{*}{\parbox{2cm}{\centering Recirculation \% Correct}} &
\multicolumn{4}{c|}{Adaptive Recirculation \% Correct}  \\ \cline{4-7}
& & & MMLU & MMLU & ARC & ARC \\ 
& & & (test) & (train) & Easy & Challenge \\ \hline
MMLU & 57.90 & \hlg{58.28} & \hlg{59.83}* & \hlg{58.63} & \hlr{56.05} & \hlr{55.02} \\ 
ARC Easy & 81.78 & \hlg{82.07} & \hlg{82.79} & \hlg{82.20} & \hlg{82.87} & \hlr{81.73} \\ 
ARC Challenge & 54.44 & \hlg{54.86} & \hlg{54.95} & \hlr{52.39} & \hlr{53.50} & \hlr{53.58} \\ 
PiQA & 79.98 & \hlg{80.52} & \hlg{80.20} & \hlg{80.30} & \hlg{80.52} & \hlg{80.09} \\ 
BoolQ & 79.08 & \hlg{79.11} & \hlg{81.04} & \hlg{79.66} & \hly{76.08} & \hlr{78.78} \\ 
WinoGrande & 69.46 & \hlr{69.38} & \hlg{69.61} & \hlr{69.38} & \hlr{67.96} & \hlr{67.96} \\ 
HellaSwag & 75.89 & \hlr{75.86} & \hlg{76.01} & \hlg{76.04} & \hlr{74.76} & \hlr{74.41} \\ 
Lambada & 70.02 & \hlg{70.27} & \hlg{71.07} & \hlr{68.58} & \hlr{62.95} & \hlr{62.22} \\ \hline
\end{tabular}
}

\vspace{1.5mm}
\caption{Accuracy of Gemma3 4B PT baseline model (second column) and model with recirculation 
(third column) on eight standard single-token response datasets.
Green shading indicates results for which recirculation matches or beats the baseline, red shading 
indicates otherwise.
Recirculation improves performance on six of eight datasets, mean 0.23\%.
The last four columns report accuracy for adaptive recirculation (Section~\ref{subsec:adaptive_recirc}), a variant of recirculation involving fine tuning of recirculation
coefficients $\alpha$ and $\beta$. The success of adaptive recirculation is highly dependent on the dataset
used for fine tuning (shown at the top of the column); consistent improvements in accuracy are observed when the MMLU test set is used
for fine tuning. * denotes overlap between the train and test distributions.}
\label{tab:single_token_benchmarks}
\end{table}

Table~\ref{tab:single_token_benchmarks} shows accuracy for eight datasets (first column)
in the baseline (second column) and recirculation (third column) models. 
Recirculation improves performance on six of the eight datasets, although the accuracy differences---both positive and negative---are modest.
The 0.23\% mean increase in accuracy is statistically reliable: with dataset as a random 
factor, a paired $t$-test indicates that a new dataset drawn from the same distribution should benefit from recirculation
($t(7)=2.80$, two-tailed $p=0.027$).
Appendix~\ref{app_sec:mc} evaluates other hyperparameter settings, finding 
reliable improvements in performance across hyperparameter settings.
For the moment, please disregard the last four columns of Table~\ref{tab:single_token_benchmarks}; we discuss them in Section~\ref{subsec:adaptive_recirc}. 

\subsubsection{GSM8k and HumanEval}

\begin{figure}[bt]
\centering
\includegraphics[width=5.5in]{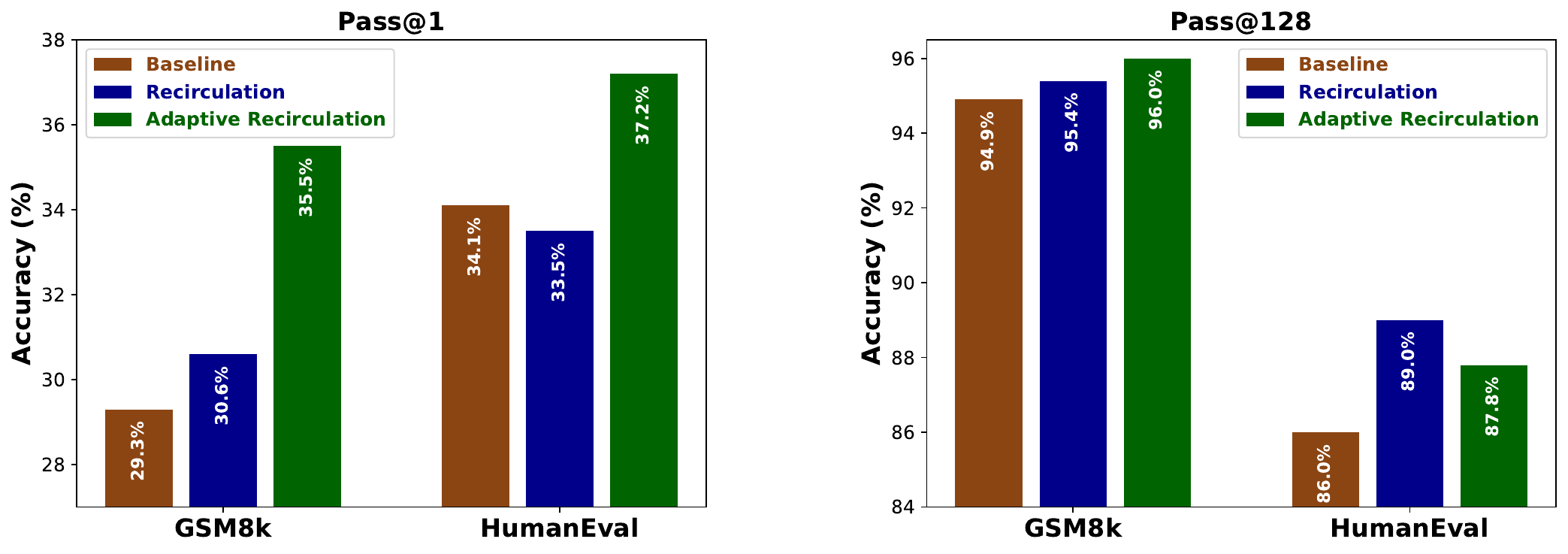}
\caption{GSM8k and HumanEval accuracy for Gemma3 4B PT with pass@1 (left panel) and pass@128 (right panel).
For each dataset, we compare the baseline (off-the-shelf) model against a version with
recirculation and a version with adaptive recirculation (see Section~\ref{subsec:adaptive_recirc}).
\label{fig:gsm_results}
}
\end{figure}

In contrast to the single-token-response datasets examined in the previous section, we turn to problems that are solved with extended reasoning \citep{kojima2022large,wei2023}, specifically the grade-school math problems in GSM8k \citep{cobbe2021gsm8k} and programming problems in HumanEval \citep{chen2021humaneval}. 
With single-token responses, any benefit of recirculation must depend on inferences made in processing the problem statement; with chain of thought and multi-token responses, recirculation has the additional opportunity to support extended reasoning.

GSM8k and HumanEval also contrast with the previous datasets in that they allow for a large set of 
candidate responses. When the number of candidates is large,
improvements in model performance can be characterized in terms of either
capability \emph{sharpening} or capability \emph{expansion} \citep{yue2025does}. The distinction
is made with pass@$k$ metrics \citep{yue2025does}: improved model performance with pass@1 (greedy decoding)
indicates sharpening---the correct response beating out other responses in
the set of candidates; improved model performance with pass@128 (accepting any of 128 samples if they are correct)
indicates expansion of the set of possibilities.
We therefore examine model performance with pass@1 (greedy decoding) and pass@128 (using a higher temperature and nucleus sampling, as recommended by \citeauthor{gemmateam2024gemmaopenmodelsbased}, \citeyear{gemmateam2024gemmaopenmodelsbased}).
Presented results are based on zero-shot evaluation.
Following \citet{kojima2022large}, for GSM8k, we use the zero-shot chain-of-thought prompt, which is important when evaluating base models without few-shot examples.

Figure~\ref{fig:gsm_results} compares baseline and recirculated Gemma3 4B models---brown and blue
bars, respectively---for GSM8k and HumanEval, with pass@1 in the left panel and pass@128 in the right. 
(We return to the green bar in  Section~\ref{subsec:adaptive_recirc}.)
Recirculation improves pass@1 performance for GSM8k but not HumanEval, though the mean improvement
across datasets is 0.4\%.  Recirculation improves pass@128 performance for both data sets,
indicating some support for both capability sharpening and capability expansion. 
Recirculation 
appears to affect extended generative responses more than single-token responses (Section~\ref{sec:mc_single_token}), 
thus revealing its potential for advanced problem solving and complex reasoning.

\subsection{Adaptive recirculation}
\label{subsec:adaptive_recirc}

\begin{figure}[b!]
    \centering
    \includegraphics[width=\linewidth]{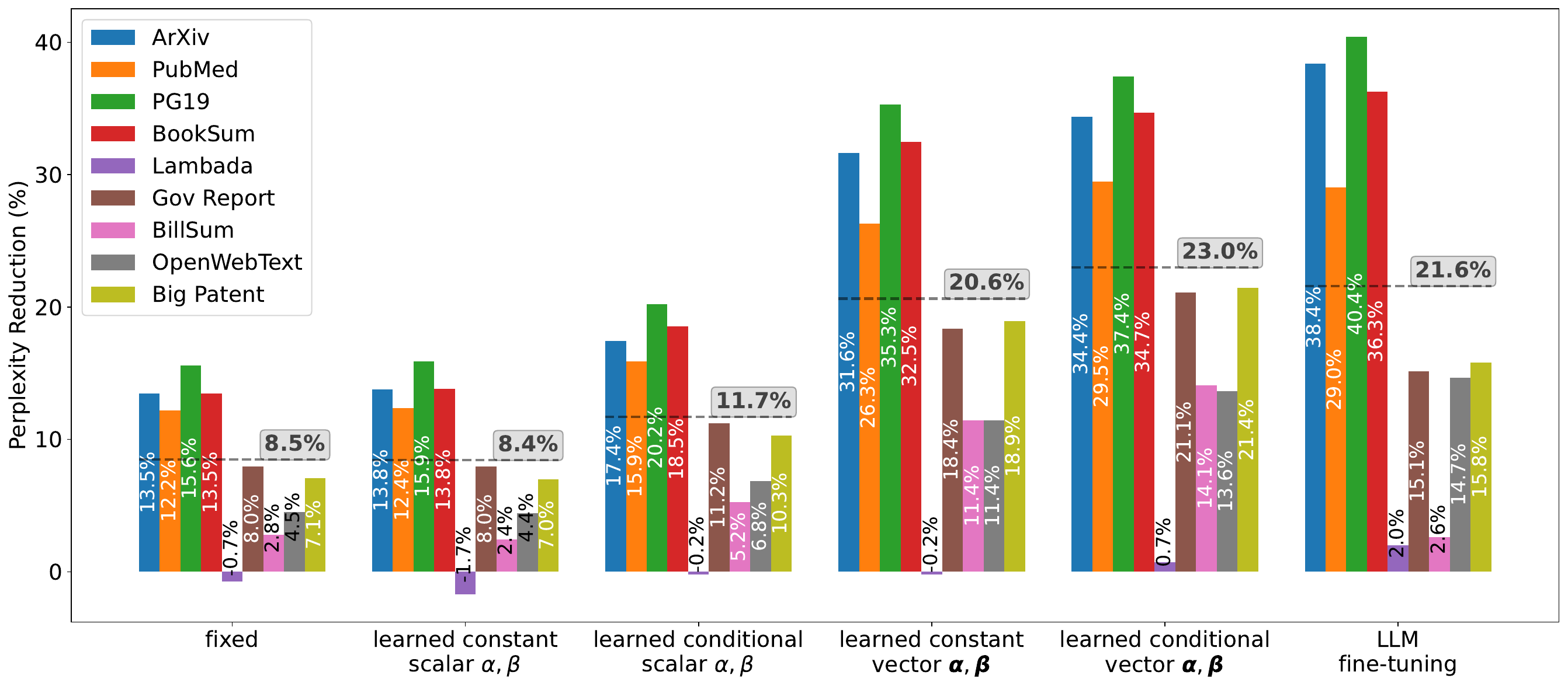}
    \caption{Reduction in perplexity (\%) with six variants of recirculation on Gemma3 1B for nine evaluation datasets. The mean percentage reduction in perplexity across datasets is shown
    via the gray dashed line above each variant.  The variant labeled `fixed' is recirculation
    with hyperparameters chosen previously. The next four variants involve learning coefficients
    $\alpha$ and $\beta$, either in a token conditional or unconditional manner and either as scalars
    or vectors. The final variant involves full fine tuning of the LLM, whose performance does
    not quite match that of the learned-conditional-vector variant, referred to in the 
    text as \emph{adaptive recirculation}.
    } \label{fig:adaptive_recirc_ablation}
\end{figure}

Our strategy to this point has been to determine how far we could push recirculation as a pure inference-time technique for
an off-the-shelf model. Recirculation is surprising and intriguing specifically because it works out-of-the-box.
The blindingly obvious next step is to improve the robustness and gains from recirculation via model adaptation. Because
training a recurrent architecture comes at a high cost and there is a risk of overfitting if the entire model is fine tuned,
we begin by taking small steps away from the training-free setting, focusing on modulating the recirculation hyperparameters 
$\alpha$ and $\beta$ (Equation~\ref{eq:recirc}). Our goal is to make changes as minimal as possible that show benefits over
the training-free recirculation setting.

For training experiments, we use Gemma3 1B PT and use a mix of documents from arXiv, C4, and PG19, and we omit BOS tokens
from many windows to amplify perplexity and make it easier to see differences among methods. 
We fix the source and destination layer indices based on our previous hyperparameter sweep.
(See Appendix~\ref{app_sec:adaptive_recirculation} for details.)
We explore methods that learn representation-mixing terms $\alpha$ and $\beta$
to minimize prediction loss. 

We score six different adaptation methods on nine datasets by the reduction in evaluation-set perplexity they achieve
relative to the baseline  (no recirculation) model. Experimental details are presented 
in Appendix~\ref{app_sec:adaptive_recirculation}. The six alternative methods, shown from left to right in Figure~\ref{fig:adaptive_recirc_ablation}, are as follows.
\begin{itemize}
    \item \textit{Fixed.} We use the fixed, previously determined coefficients ($\alpha=0.15$, $\beta=0.85$),
    thereby replicating earlier results.

    \item \textit{Learned constant scalars}. We perform gradient descent in the scalars $\alpha$ and $\beta$
    to best fit the training data.

    \item \textit{Learned conditional scalars}. We train an MLP that maps the token-specific source and
    destination embeddings to scalar $\alpha$ and $\beta$ values, allowing the model to self-determine
    the strength of recirculation for each token.

    \item \textit{Learned constant vectors.} Instead of scaling the source and destination vectors by 
    constants when recirculating, we scale by learned vector-valued $\boldsymbol{\alpha}$ 
    and $\boldsymbol{\beta}$:
    \begin{equation}
    \bz{t+1,t,d} = \boldsymbol{\alpha} \circ f(\bz{t,t,s}) + \boldsymbol{\beta} \circ \bz{t,t,d} ,
    \label{eq:recircvec}
    \end{equation}
    where $\circ$ is the Hadamard product.

    \item \textit{Learned conditional vectors.} We train an MLP that maps the token-specific source and 
    destination embeddings to vector-valued $\boldsymbol{\alpha}$ and $\boldsymbol{\beta}$.

    \item \textit{Model fine tuning.} We fine tune the Gemma3 1B architecture augmented with the
    recurrent connections of recirculation. We fix the mixture coefficients at $\alpha=0.15$ and $\beta=0.85$.
\end{itemize}

Of the methods that learn coefficients $\alpha$ and $\beta$, 
the two that learn vector-valued coefficients (fourth and fifth sets) perform better than those that learn scalar 
coefficients (second and third sets). And the two methods that learn token-conditional coefficients (third and fifth set)
outperform the two methods that learn static coefficients (second and fourth set). 
Of the four methods, the best (fifth set: learned conditional vector $\boldsymbol{\alpha}, \boldsymbol{\beta}$)
trains an MLP to generate token-specific mixture vectors.
The other methods (second, third, and fourth sets) can be considered ablations demonstrating that both
vector-valued coefficients and conditioning on the current token is essential.
Henceforth, we refer to the superior method as \emph{adaptive recirculation}.

Adaptive recirculation obtains a mean 23.0\% reduction in perplexity, relative to the 8.5\% reduction for basic recirculation.
Adaptive recirculation beats recirculation for each of the nine datasets.
Adaptive recirculation performs at least as well as full Gemma3 1B fine tuning (23.0\% versus 21.6\%).
The advantage of adaptive recirculation is that the Gemma3 model itself is untouched and therefore there is
less risk of overfitting the model via fine tuning.

On downstream tasks, adaptive recirculation also shows promise, with the caveat that the dataset used for training the MLP
is critical.
Returning to the eight single-token-response datasets, the last four columns of Table~\ref{tab:single_token_benchmarks} present accuracy for adaptive recirculation on these evaluation datasets for four different training datasets.
Adapting to a small fraction of the MMLU test split (3600 of about 14000 examples) yields robust across-the-board improvements, and adapting to MMLU (auxiliary train split) yields on balance improvements.
For comparability, we use the full test set of MMLU for evaluation even when training on the MMLU test split, which includes the examples used for training; performance on other datasets highlights generalization.
Note that the auxiliary train set for MMLU is in general much lower quality than the well-curated MMLU test set, which may be the reason for its inferior performance.
Finally, the MMLU train set is based on a combination of existing datasets, with potential contamination \citep{hendryckstest2021}.
Training directly on other datasets such as ARC Easy or ARC Challenge produces significant drops in performance, although these
datasets have a much smaller number of examples.
Details of these experiments are provided in
Appendix~\ref{app_sec:adaptive_recirculation}.

Although neither recirculation nor adaptive recirculation yields robust accuracy gains on
single-token response tasks, the gains for GSM8k and HumanEval offer a promising signal for extended generative 
response tasks. The green bar in each panel of Figure~\ref{fig:gsm_results} shows GSM8k and HumanEval performance
with adaptive recirculation. Although the original model parameters are untouched, adaptive recirculation 
yields 8.8\% (pass@1) and 20.9\%  (pass@128) reductions in error rate for GSM8k, and
4.7\% (pass@1) and 12.9\% (pass@128) reductions in error rate for HumanEval.

\section{Discussion}

Our work explores the consequences of recirculation, a training-free, inference-time architectural modification
to incorporate recurrence into a transformer for improved state tracking. We have shown that for the Gemma3
pretrained and instruction-tuned architectures, recirculation yields notable reductions in model perplexity 
and enhances models' ability to follow instructions, answer questions, and problem solve. 
When generating responses, recirculation incurs almost no additional computation cost; although recirculation 
requires running two transformer stacks in parallel instead of a single stack, modern AI hardware parallelizes 
efficiently in this case. However, there is an additional cost of processing the prefill context autoregressively, 
which for large-context problems, can be quite slow.

Although there is a robust literature on architectural improvements to the transformer, this literature focuses almost
exclusively on training models from scratch or incorporating modifications mid-training. The notable exception
is the notion of training-free looping \citep{li2025skiplayerloopit,chen2026,ng2026rys}, a related but 
distinct architectural modification. Both looping and recirculation exploit a key property of the transformer's
residual pathway:  the alignment of representations across layers \citep{elhage2021mathematical}.
Recirculation is complementary to and can be combined with other methods in common practice, including looping,
variable computation time, and the coarser-grain recurrence that occurs with both latent thought and
chain of thought.  

\subsection{Model-design affordances}

We view recirculation not as a `shovel ready' technique to be incorporated into
state-of-the-art models, but rather as a methodological or philosophical
contribution. By focusing on training-free architectural modifications, we
are essentially asking the model to reveal to us its ingrained paths for improvement.
Our search over where to place recurrence in the architecture and how to mix and
normalize activations is informed by the model itself.
The alternative---the typical course of research in machine learning---is to 
propose an arbitrary modification to an architecture and then evaluate the potential
of this modification via costly training. Our key hypothesis is that if we 
identify how the pretrained model \emph{wants} us to tweak it without modifying its weights,
we also identify an inductive bias that will facilitate and simplify training, whether
from scratch or with fine tuning. Our adaptive recirculation experiments support
this hypothesis.

In product design, practitioners talk about design \emph{affordances} \citep{norman1999},
which are properties of an object that inform us about how it should be used.
For example, a door handle affords grasping and pulling; a door plate affords
pressing. Analogously, our research explores \emph{model-design affordances}---the 
natural properties of a foundation model that we can exploit and amplify to improve the
model's basic operation. By improving this basic substrate, we improve all capabilities
that build on it. Most approaches to incorporating recurrence within the
internal layers of a foundation model do so via adapters or cross attention. Our
observations suggest that the simpler strategy of mixing activation vectors may suffice.

\subsection{Limitations and future directions}

We close by discussing limitations of the present work and promising future directions that will address these
limitations.

\begin{itemize}[leftmargin=*]
\item
\textit{Determination of optimal hyperparameters.}
The optimal model hyperparameters---the source and destination layers and the recirculation coefficients
$\alpha$ and $\beta$---appear to be domain or problem dependent. 
However, hyperparameters selected based on one dataset or one criterion often generalize to another. For example,
hyperparameters based on perplexity evaluations were used for downstream tasks; hyperparameters based 
on one specific dataset (MMLU) were used for a range of other single-token-response tasks; and small sets of 
examples are often sufficient to select hyperparameters. However, unless
we can identify task-universal hyperparameters or an automatic mapping from task
to hyperparameters, the practical applicability of recirculation will be limited.

\item
\textit{Normalization.} We find that normalization of source activation vectors makes recirculation behave
more robustly. For the Gemma3 family, we find that of the many candidate normalization schemes we investigated, 
one in particular seems to work well across model sizes (Section~\ref{app_sec:normalization_ramping}). Nonetheless, 
further experiments are needed to determine whether normalization is necessary and the best normalization scheme for
other model families.

\item
\textit{Blockwise recurrence.} Recirculation is computationally cheap during autoregressive generation because
running two stacks in parallel is nearly as efficient with modern hardware as one stack. However, during prefill,
token-by-token processing is also required. For state-of-the-art foundation models, which may include a large
amount of contextually relevant information, sequential processing of the prefill may be infeasible.
One way to alleviate the computational burden would be to perform blockwise recirculation, i.e., to process a set of $K$
tokens in parallel, and then recirculate those $K$ tokens simultaneously with the next $K$ as they are processed for
the first pass. We have yet to explore the trade off between $K$ and recirculation performance.

\item \textit{Multiple recirculation iterations.}  
A true recurrent network would recirculate the activation in stacks 1 to $t-1$ at step $t$. In contrast, the variant of recirculation
we explored recirculates only stack $t-1$, such that each stack is recirculated exactly one time. We have yet to study an intermediate
approach in which each stack is recirculated exactly $r$ times. Figure~\ref{app_fig:recirc_2iterations} shows an unrolled architecture
with $r=2$ iterations per stack. We wished to start with the simplest instantiation, but given the low cost of recirculating $r$ stacks
in parallel, increasing the number of iterations is worthwhile.

\item \textit{Multiple recirculation paths.}
We picked a single \{source, destination\} pair for recirculation, but one could recirculate along multiple paths simultaneously.
Indeed, for some architectures (e.g., see Gemma3 4B and 12B recirculation sweeps in Figure~\ref{fig:looping_vs_recirc}), there 
appeared to be distinct, disjoint regions of the hyperparameter sweep where recirculation was effective. One hypothesis is that
these distinct regions are complementary, e.g., they may convey state information at different levels of abstraction.

\item \textit{Adaptation.}
The present work has only scratched the surface of possible approaches to adapting models to enhance recirculation effects.
Our results with adaptive recirculation are extremely promising considering that the tuning set we used was quite small and that 
tuning of model weights was not required to obtain robust performance improvements. Before turning to model-weight fine tuning,
many avenues for improvement still exist, e.g., token conditioned recirculation pathways and normalization methods.
\end{itemize}

By listening to the model's own internal dynamics rather than forcing costly architectural overhauls, recirculation offers a powerful, computationally inexpensive path forward. Unlocking these intrinsic affordances may solidify the model's basic contextual understanding, which in turn provides the vital scaffolding required for extended, multi-turn reasoning.
\section{Acknowledgements}
Our thanks to Lei Bao, Mukul Bhutani, and Lan Le who contributed to the shaping of this research. We are also grateful to Michael Riegler and Klas Pettersen (SimulaMet, Norway), Patrick Haller (Humboldt U., Berlin), Yang Wang (Microsoft), and Jonathan Yanay for their independent replications and constructive discussions of our results.
\bibliography{main}

\newpage

\appendix

\renewcommand{\thetable}{\thesection.\arabic{table}}
\renewcommand{\thefigure}{\thesection.\arabic{figure}}

\section{Unrolled recirculation architecture}
\setcounter{table}{0} 
\setcounter{figure}{0} 
\label{app_sec:addl_fig}
Figure~\ref{app_fig:recirc_2iterations} shows an unrolled transformer with two iterations of recirculation.  In general,
with $k$ iterations of recirculation, it will be necessary to run $k+1$ stacks for each input step.

\begin{figure}[htb]
    \centering
    \includegraphics[width=5.5in]{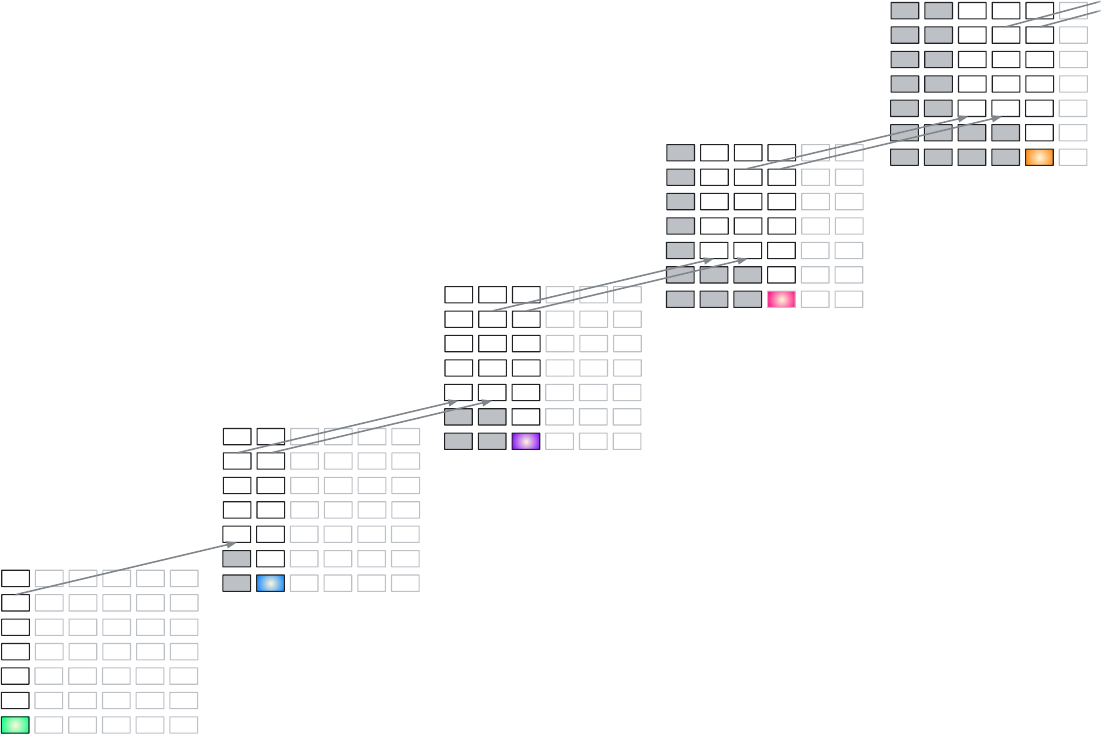}
    \caption{Recirculation architecture with two iterations of recirculation, which yields three passes through each stack.
    } \label{app_fig:recirc_2iterations}
\end{figure}

\newpage
\section{Recirculation implementation details}
\setcounter{table}{0} 
\setcounter{figure}{0} 
\subsection{Hyperparameter sweeps}
\label{app_sec:hyperparm_sweeps}
For the hyperparameter sweeps (Figures~\ref{fig:hyperparm_sweep} and \ref{fig:hyperparm_sweep_3combined}), we used a context
window of 1024 tokens and processed documents from the designated training sets of arXiv, C4, and PG19.  Table~\ref{app_tab:sweep_dataset_stats}
shows the number of documents and tokens evaluated from each dataset.
For the final window of each document, we discarded the window if fewer
than 32 tokens were present and filler tokens were appended as
necessary but were not included in the perplexity calculation.  A beginning-of-sequence (BOS) token was inserted at
the start of each window, as Gemma expects.
\begin{table}[b!]
\centering
\begin{tabular}{lrr}
\toprule
Dataset & Documents & Total Tokens \\
\midrule
arxiv   & 100       & 923,769      \\
C4      & 2000      & 817,377      \\
PG19    & 10        & 1,137,576    \\
\bottomrule
\end{tabular}
\caption{Sweep Data Statistics}
\label{app_tab:sweep_dataset_stats}
\end{table}
\begin{figure}[b!]
    \centering
    \includegraphics[width=1.0\linewidth]{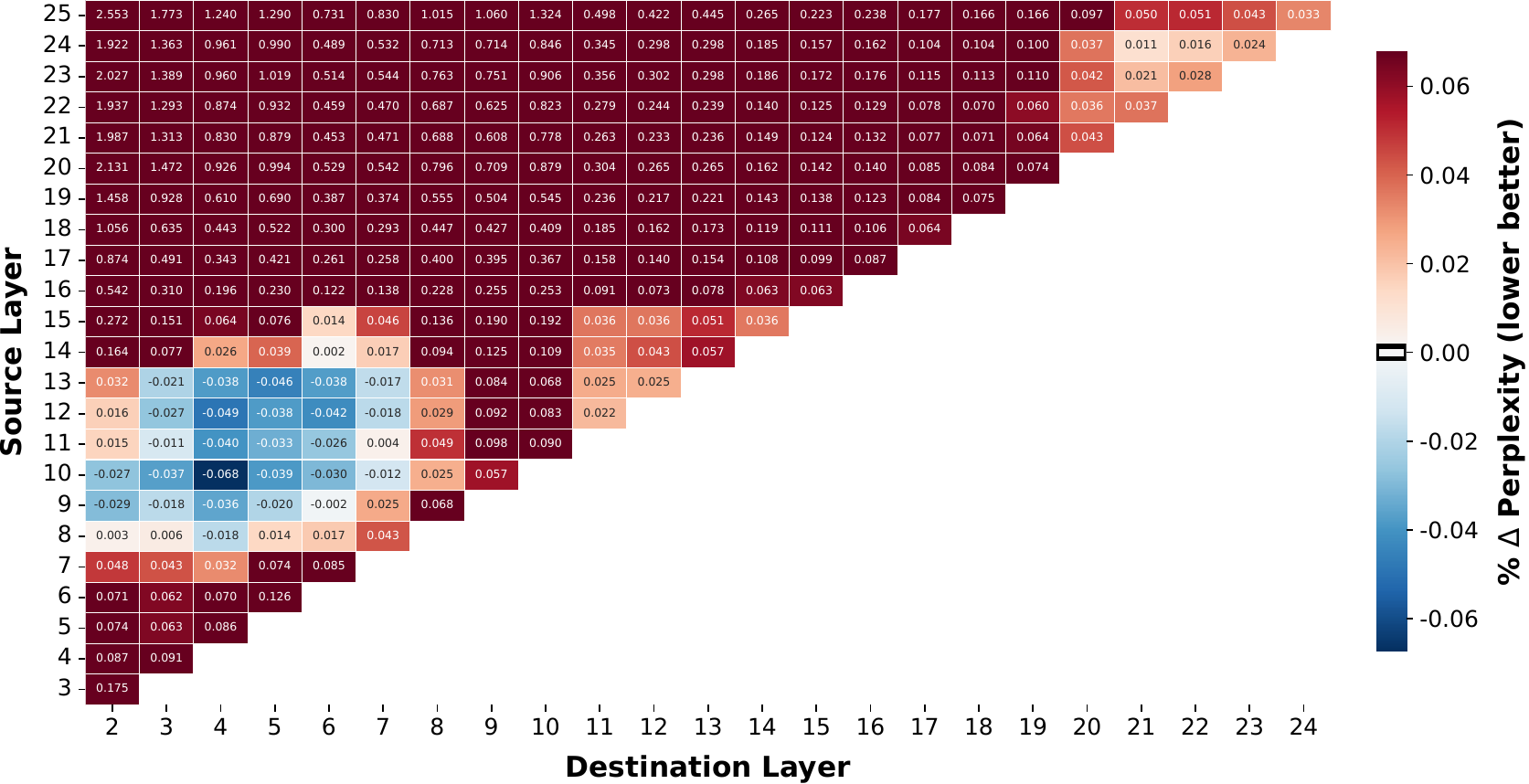}
    \caption{Source and destination layer sweep of Gemma3 1B PT with recirculation and
    $\alpha=0.05$.}
    \label{app_fig:sweep_1b}
\end{figure}
\begin{figure}[tb]
    \centering
    \includegraphics[width=1.0\linewidth]{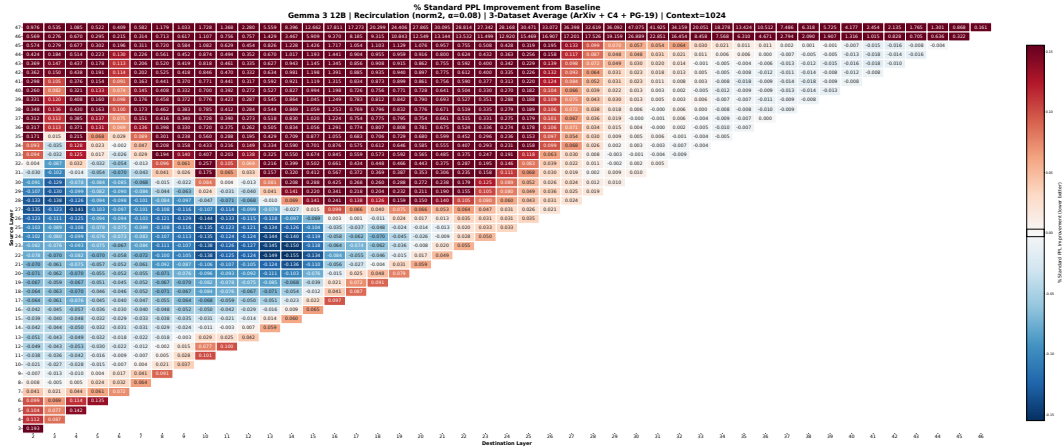}
    \caption{Source and destination layer sweep of Gemma3 12B PT with recirculation and
    $\alpha=0.08$.}
    \label{app_fig:sweep_12b}
\end{figure}

Source and destination sweeps for the 1B and 12B PT models
are shown in Figure~\ref{app_fig:sweep_1b} and \ref{app_fig:sweep_12b},
respectively. We use $\alpha=.05$ for the 1B and $\alpha=.08$ for the 12B, as determined by a more systematic sweep 
over all three hyperparameters.
Based on hyperparameter sweeps for the 1B, 4B, and 12B PT models, we selected the source and destination
pairs shown in Table~\ref{app_tab:hyperparameters}. These hyperparameters were used for all 
results presented in the article unless otherwise mentioned.
We also conducted sweeps for BOS-absent sequences, which yields the hyperparameters also shown in Table~\ref{app_tab:hyperparameters}. Some of our experiments prefer BOS-absent sequences to amplify effects of recirculation and better compare among alternative methods.

\begin{figure}[b!]
    \centering
    \includegraphics[width=1.0\linewidth]{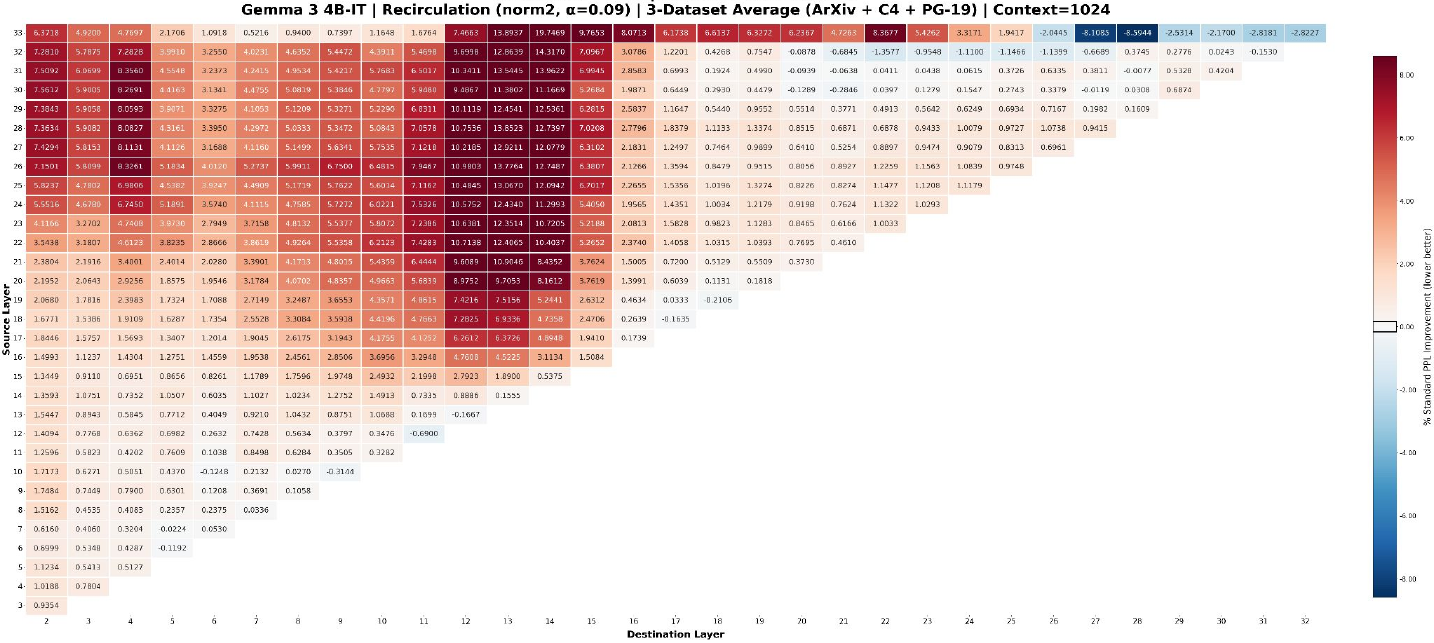}
    \caption{Source and destination layer sweep of Gemma3 4B IT model, with recirculation and $\alpha=0.09$.}
    \label{app_fig:gemma3_4b_it_sweep}
\end{figure}
\begin{figure}[bt]
    \centering
    \includegraphics[width=1.0\linewidth]{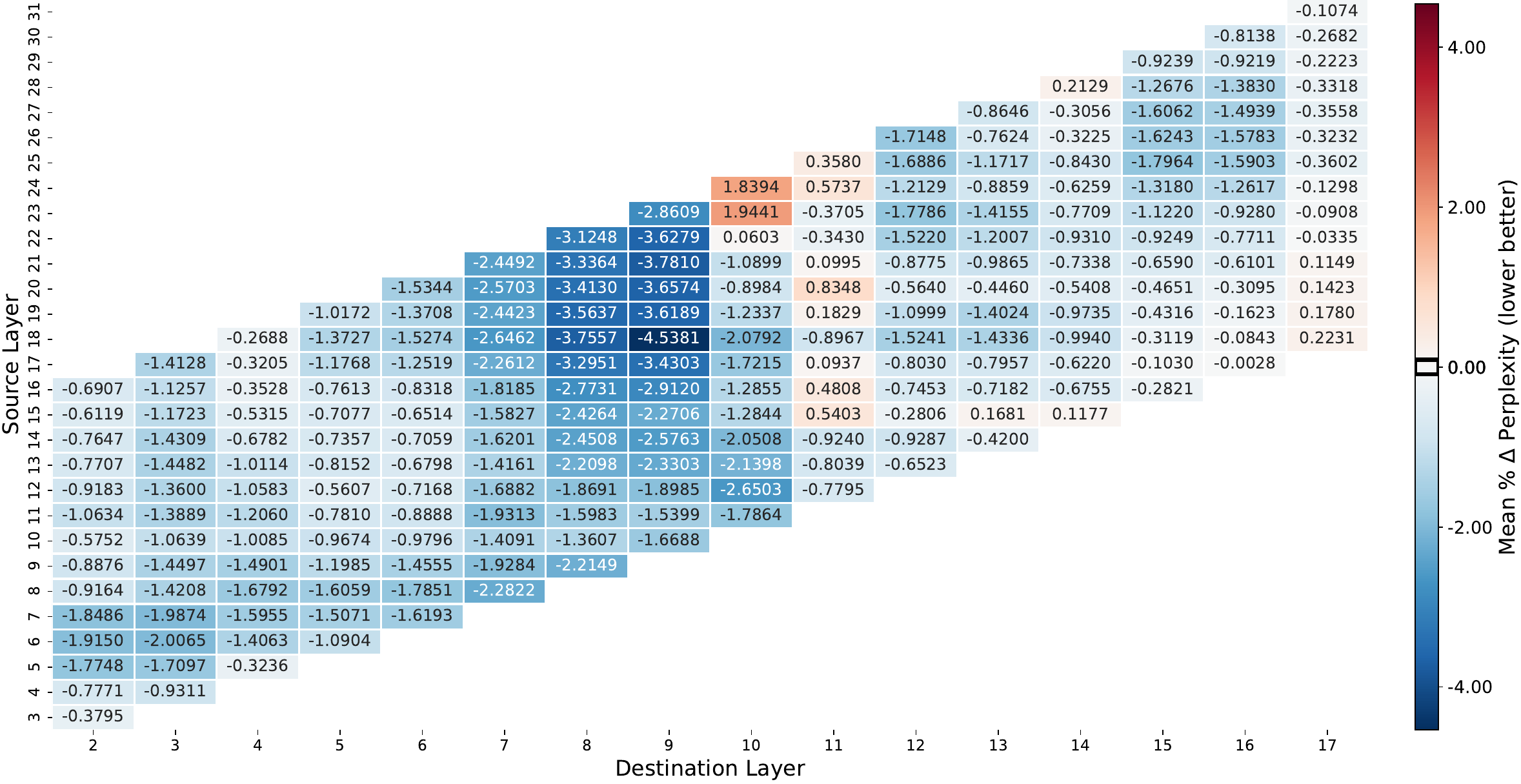}
    \caption{Source and destination layer sweep of Gemma3 4B PT with recirculation and
    $\alpha=0.10$, where the beginning-of-sequence (BOS) token is absent from half the sequences.}
    \label{app_fig:gemma3_4b_bos_sweep}
\end{figure}

For the Gemma3 4B IT model, we conducted an independent sweep (Figure~\ref{app_fig:gemma3_4b_it_sweep}) that identifies 
source layer 33, destination layer 28, and $\alpha = 0.15$.
For the experiments with the BOS (beginning of sequence) corrupted windows, we conducted a sweep
with the pretrained model (Figure~\ref{app_fig:gemma3_4b_bos_sweep}) that identifies source layer 18, destination layer 9, 
and $\alpha=0.15$; the source and destination are not far off from those obtained with in-distribution sequences (source 15,
destination 8).

\begin{table}[bt!]
\centering
\begin{tabular}{llccccc}
\toprule
Model & Condition & Source & Destination & $\alpha$ & $\beta$ & Ramp Steps\\
\midrule
\multirow{2}{*}{Gemma3 1B} & PT & 10 & 4 & 0.07 & 1.00 &  \multirow{2}{*}{10} \\
& PT, BOS absent & 11 & 4 & 0.07 & 0.93 & \\
\hline
\multirow{3}{*}{Gemma3 4B} & PT  & 15 & 8 & 0.09 & \multirow{3}{*}{1.00} & \multirow{3}{*}{0} \\
& PT, BOS absent  & 18 & 9 & 0.10 & & 0 \\
& IT & 33 & 28 & 0.15 & & \\
& PT, MMLU devel. set & 16 & 5 & 0.09 & & \\
\hline
\multirow{2}{*}{Gemma3 12B} & PT & 22 & 13 & 0.08 & \multirow{2}{*}{1.00}  & \multirow{2}{*}{0} \\
& PT, BOS absent & 35 & 16 & 0.10 & & \\
\bottomrule
\end{tabular}
\caption{Optimal hyperparameters found by a sweep. Other than the row labeled ``MMLU tuning set,'' all sweeps used subsets of arXiv, PG-19, and C4,
as described in Appendix~\ref{app_sec:hyperparm_sweeps}.  PT~=~pretrained model, IT~=~instruction tuned model, BOS absent~=~only first window
in a document included the BOS token.}
\label{app_tab:hyperparameters}
\end{table}

\subsection{Perplexity evaluation}
\label{app_sec:perplexity_evaluation}

For our perplexity evaluation (Table~\ref{tab:recirc_ppl}), we used the entire evaluation set from eight datasets (arXiv, BillSum, BookSum/chapter, C4/webtextlike, Gov Report, LAMBADA, PG-19, and PubMed), and the first 10,000 documents from two larger corpora (Big Patent and Newsroom). The evaluation split was labeled `validation' for C4 and `test' for all other datasets. We partitioned each document into consecutive windows of 1024 tokens, retaining all partial windows with at least 32 tokens and masking padding tokens from the loss calculation. This procedure resulted in the evaluation of the number of tokens listed in the second column of Table~\ref{tab:recirc_ppl}.

Table~\ref{app_tab:perplexity} evaluates the same data sets (except using booksum/book instead of booksum/chapter) with
windows in which the BOS token is omitted from all but the initial window of a document. Results are based on the optimal
hyperparameters chosen for this condition (Table~\ref{app_tab:hyperparameters}). For these sweeps, only full windows
of tokens were used, i.e., we discarded windows with filler.
\begin{table}[tbp]
\centering
\renewcommand{\arraystretch}{1.3}
\resizebox{\textwidth}{!}{
\begin{tabular}{|l|r|rrr|rrr|rrr|}
\hline
\multicolumn{1}{|c|}{\textbf{Dataset}} & \textbf{\# tokens} & \multicolumn{3}{c|}{\textbf{Gemma3 1B PT}} & \multicolumn{3}{c|}{\textbf{Gemma3 4B PT}} & \multicolumn{3}{c|}{\textbf{Gemma3 12B PT}} \\ \cline{3-11} 
\multicolumn{1}{|c|}{} & & \textbf{baseline ppl} & \textbf{recirc. ppl} & \textbf{\% reduction} & \textbf{baseline ppl} & \textbf{recirc. ppl} & \textbf{\% reduction} & \textbf{baseline ppl} & \textbf{recirc. ppl} & \textbf{\% reduction} \\ \hline
\textbf{arxiv} & 51M & 19.10 & 16.54 & \hlg{13.99\%} & 14.76 & 13.10 & \hlg{11.26\%} & 33.93 & 25.38 & \hlg{25.20\%} \\ 
\textbf{big\_patent} & 55M & 10.89 & 9.90 & \hlg{9.13\%} & 8.74 & 8.12 & \hlg{7.08\%} & 17.55 & 13.20 & \hlg{24.76\%} \\ 
\textbf{billsum} & 5.8M & 4.71 & 4.65 & \hlg{1.25\%} & 3.57 & 3.41 & \hlg{4.46\%} & 4.50 & 4.09 & \hlg{10.96\%} \\ 
\textbf{booksum/book} & 7.7M & 32.48 & 27.30 & \hlg{15.95\%} & 29.09 & 24.45 & \hlg{15.95\%} & 77.02 & 51.67 & \hlg{32.91\%} \\ 
\textbf{c4/webtextlike} & 4M & 17.10 & 16.43 & \hlg{3.93\%} & 14.26 & 13.64 & \hlg{4.37\%} & 19.21 & 16.71 & \hlg{13.01\%} \\ 
\textbf{gov\_report} & 3.4M & 11.86 & 11.00 & \hlg{7.26\%} & 10.79 & 9.78 & \hlg{9.37\%} & 27.59 & 19.11 & \hlg{30.74\%} \\ 
\textbf{lambada} & 418k & 35.62 & 35.88 & \hlr{$-$0.72\%} & 28.34 & 28.19 & \hlg{0.53\%} & 25.47 & 26.19 & \hlr{$-$2.81\%} \\ 
\textbf{newsroom} & 95M & 14.11 & 13.75 & \hlg{2.56\%} & 11.62 & 11.17 & \hlg{3.87\%} & 13.74 & 12.45 & \hlg{9.40\%} \\ 
\textbf{pg19} & 10.8M & 22.27 & 19.06 & \hlg{14.41\%} & 19.49 & 16.43 & \hlg{15.72\%} & 52.86 & 34.15 & \hlg{35.40\%} \\ 
\textbf{pubmed} & 23M & 15.73 & 13.99 & \hlg{11.08\%} & 11.65 & 10.58 & \hlg{9.21\%} & 26.38 & 19.92 & \hlg{24.49\%} \\ \hline
\end{tabular}
}
\vspace{1.5mm}
\caption{Reduction in perplexity across ten language-modeling tasks with recirculation for Gemma3 models when BOS 
tokens are omitted.}
\label{app_tab:perplexity}
\end{table}

\subsection{Normalization and ramping} 
\label{app_sec:normalization_ramping}
\begin{figure}[tb]
    \centering
    \includegraphics[width=5.5in]{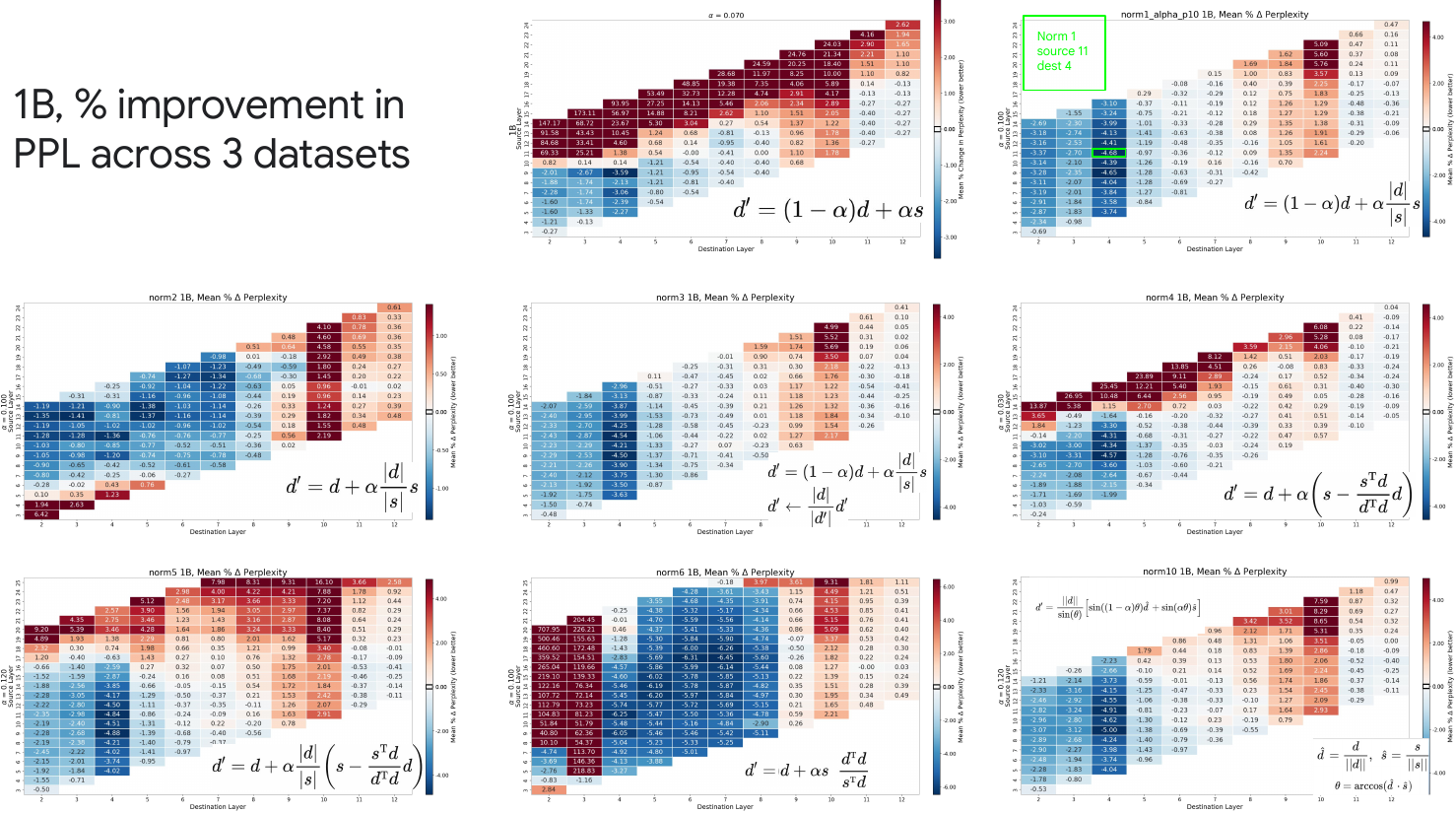}
    \caption{Hyperparameter sweeps for various normalization schemes with Gemma3 1B PT.}
    \label{fig:normschemes1b}
\end{figure}
\begin{figure}[tb]
    \centering
    \includegraphics[width=5.5in]{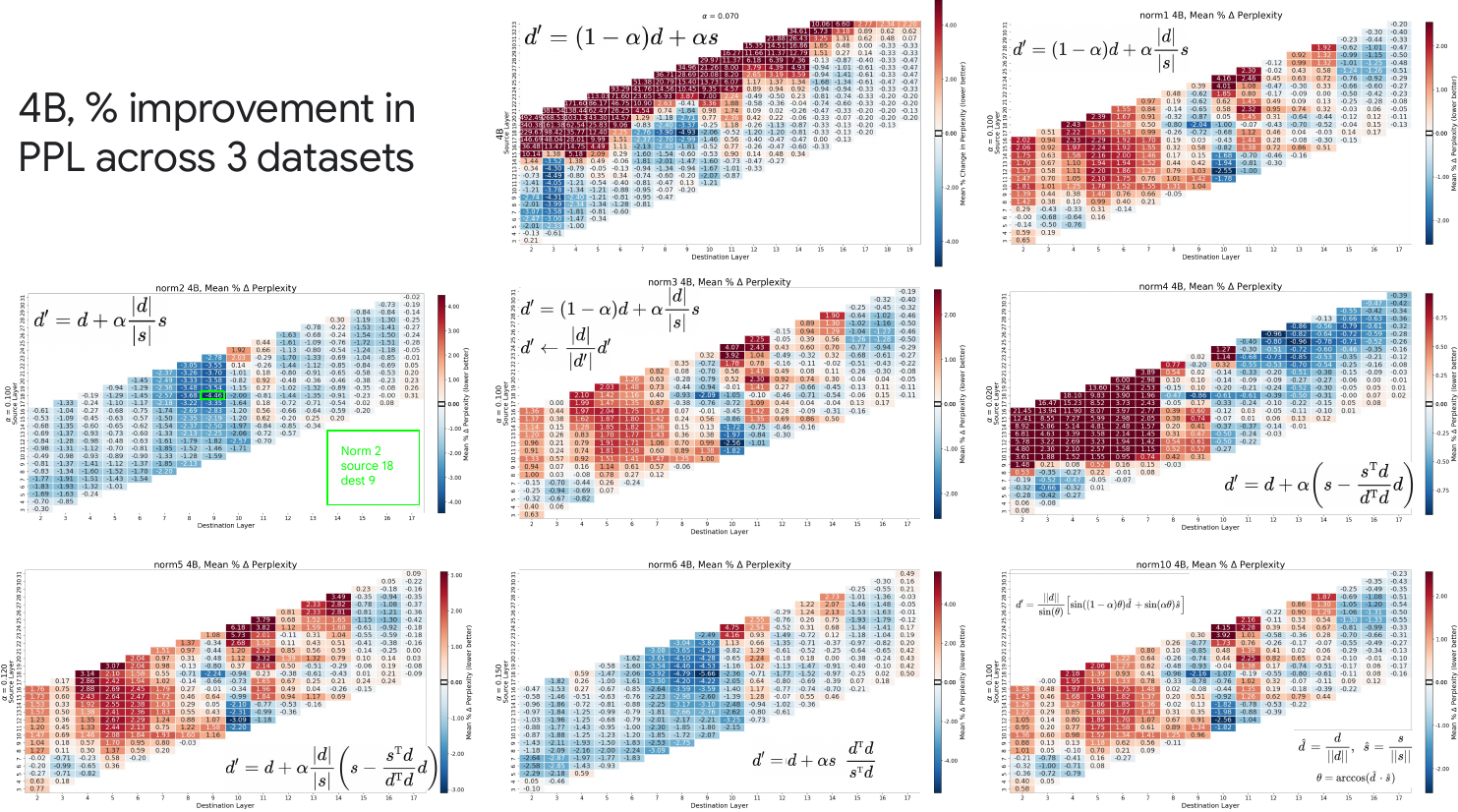}
    \caption{Hyperparameter sweeps for various normalization schemes with Gemma3 4B PT.}
    \label{fig:normschemes4b}
\end{figure}
\begin{figure}[tb]
    \centering
    \includegraphics[width=5.5in]{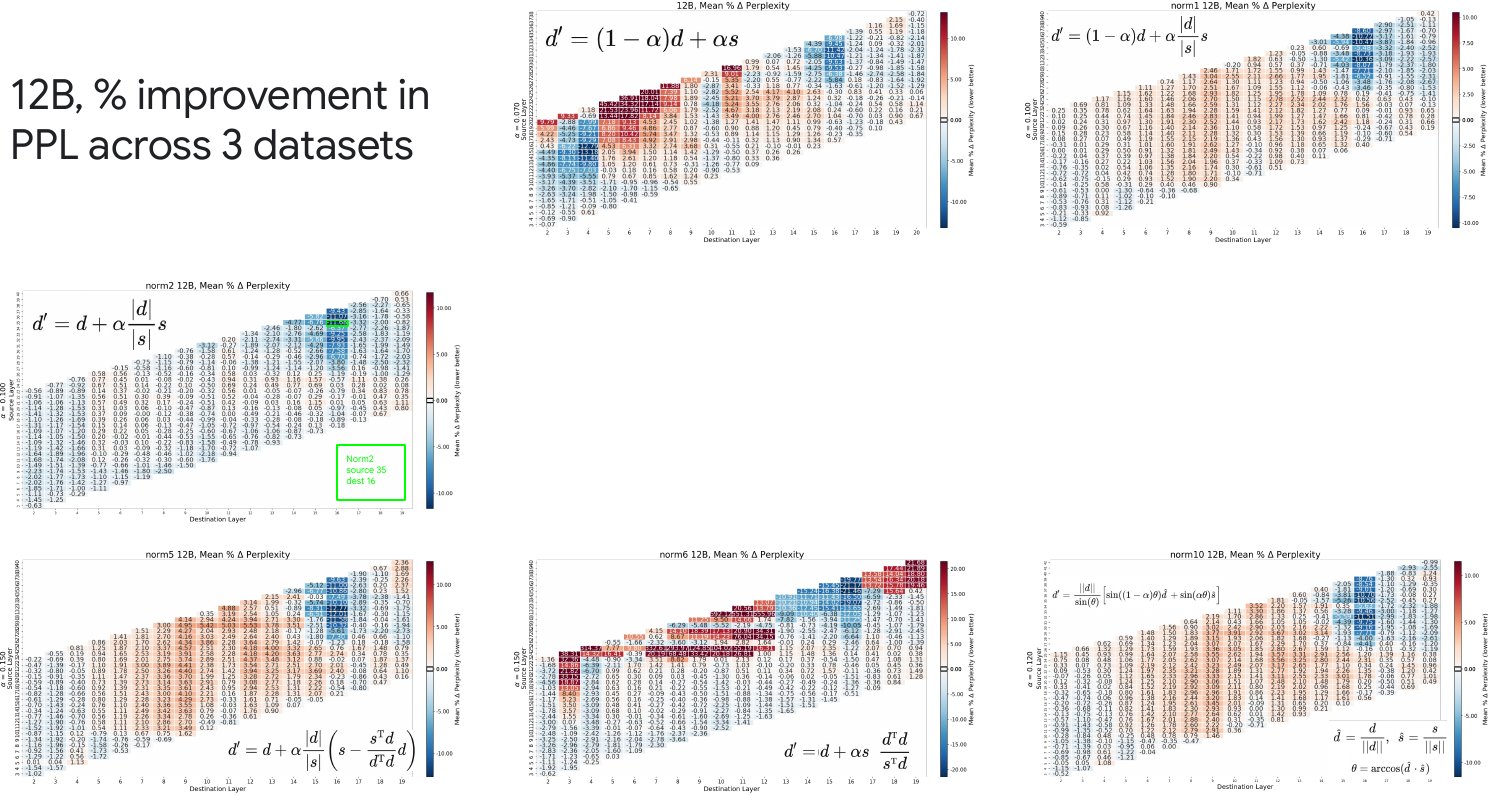}
    \caption{Hyperparameter sweeps for various normalization schemes with Gemma3 12B PT.}
    \label{fig:normschemes12b}
\end{figure}

Figures~\ref{fig:normschemes1b}-\ref{fig:normschemes12b} show hyperparameter sweeps for multiple renormalization
techniques, i.e., scaling of the source vector when recirculating to the destination layer.
Scaling helps to ensure that the source activations do not overwhelm the destination activations, due to the fact that 
embedding norms increase over transformer layers \citep{Liu2020,Xiong2020}. 
The heatmaps depict percentage reduction (blue) or increase (red) in perplexity over our three tuning datasets (arXiv,
C4, PG-19). For these heatmaps, we use two windows from each document and include a BOS token only for the first window,
which magnifies the perplexity-reduction benefits of recirculation and makes it easier to visualize similarities among
normalization schemes.

\include{recirculation_equations.tex}
The normalization schemes are summarized in the Figures and in Table~\ref{app_tab:normalization_schemes}.
We simplify the notation used in the main paper, where $\boldsymbol{z}_{i,j,l}$ referred to the embedding at 
unrolling step $i$ for input step $j$ and layer $l$. Instead, we denote:
\begin{align*}
d \equiv~ & \boldsymbol{z}_{t,t,d},  \\
s \equiv~ & \boldsymbol{z}_{t,t,s}\text{, and} \\
d' \equiv~ & \boldsymbol{z}_{t+1,t,d}.
\end{align*}
Each scheme is expressed in terms of a mixture of the destination and source activation vectors, $d$ and $s$, as well as in terms of $d$ and $s_{\perp}$, the component of $s$ orthogonal to $d$.
The normalization schemes in Table~\ref{app_tab:normalization_schemes} include the simple scheme in which no 
normalization is applied.  The no-normalization heatmaps (top row, middle panel in 
Figures~\ref{fig:normschemes1b}-\ref{fig:normschemes12b}) attain reasonable outcomes but clearly the schemes with
$L_2$ normalization (top row, right panel and second row, left panel) are better behaved in the sense that there are fewer
hyperparameters that result in poorer performance. Oddly, a convex combination of source and destination vectors 
(top row, right panel) is superior to a nonconvex combination (second row, left panel) for the Gemma3 1B model,
but the nonconvex combination is superior for Gemma3 4B and 12B. Other candidate schemes either had pathologies
(non-smooth heatmaps, large red regions) or mimicked the simple norm-ratio schemes. We have not considered normalization
of other model families and thus have no recommendations, but the heatmaps in Figure~\ref{fig:other_architectures} use
the norm-ratio scheme.

\textit{Ramping.} As we mentioned in the main article, for the Gemma3 1B model, we found a small additional reduction in
perplexity if we \emph{ramped} up the recirculation coefficient over the first $\rho=10$ steps. Specifically, we defined the $\alpha$ coefficient
at step $t \ge 0$  to be $\alpha_t \equiv \min (t/\rho,1) \alpha$.


\section{Basic results}
\setcounter{table}{0} 
\setcounter{figure}{0} 
\subsection{Robustness across model families}
\label{app_sec:architecture_robustness}

In Figure~\ref{fig:other_architectures} of the main article, we show recirculation hyperparameter sweeps for
four model families: Ministral3, Qwen3, Pythia, and Phi2. The Figure compares
architectures using only the arXiv training set (with BOS tokens). We fix $\alpha=0.07$ and $\beta=1.0$.

\begin{figure}[!b]
    \centering
    \includegraphics[width=5.5in]{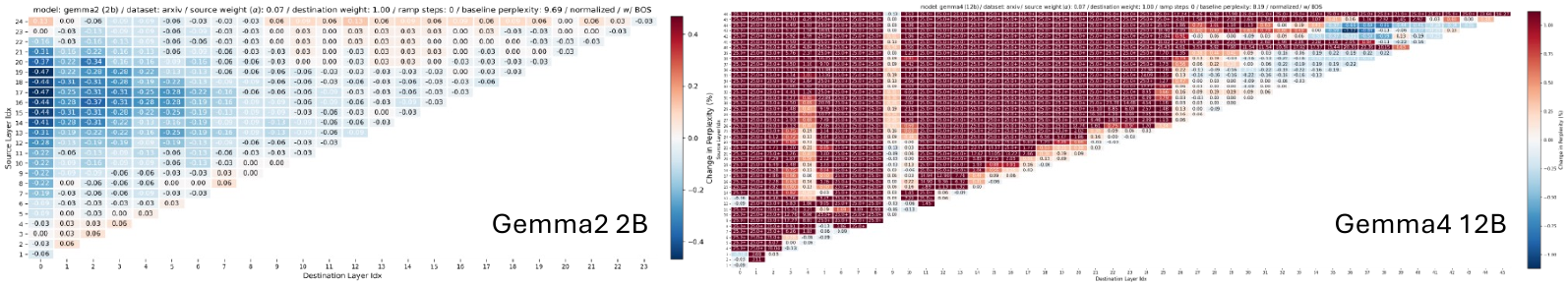}
    \caption{Recirculation hyperparameter sweeps on Gemma2 2B and Gemma4 12B on a subset of the arXiv dataset with $\alpha=0.07$.
    } \label{app_fig:other_architectures_gemma}
\end{figure}

In addition to the comparison across different model families, we also compared results from Gemma3
with those from other generations of the Gemma model family,
including the older Gemma2 2B and the newer Gemma4 12B (Figure~\ref{app_fig:other_architectures_gemma}).
The arXiv train set is again used in this Figure. While the heatmap pattern for Gemma2 matches that for Gemma3 and other model
families, Gemma4 appears to show quite a different pattern. However, the difference arises due to the fact that there's a deep 
source-destination pair (layers 43 and 36, respectively) where recirculation yields very large effects, making the smaller effects
less visible on the heatmap. Note the narrow region early in the model's processing sequence shows typical recirculation effects.
The late-stage recirculation that Gemma4 seems to be calling for may be related to a similar pattern we found with instruction-tuned
models, where the value of recirculation was to soften the output distribution (see middle panel of Figure~\ref{app_fig:tempsweep}).
Note that it may not be advisable to try recirculation for smaller variants of Gemma4 (i.e., E2B and E4B) that use cross-layer KV 
cache sharing and per-layer embedding  \citep{team2026gemma}.

The universality of an admissible region of (source, destination) pairs across all model families suggests the robustness
of the phenomenon given architectural variations and differences in the training procedure.  For example,
Gemma architectures use a \textit{Peri-LN} architecture \citep{kim2025peri} which adds layer-norm to both the input of the layer (i.e., self-attention or MLP), as well as the output of the layer. In contrast, other architectures only use input normalization \citep{Xiong2020}. 

\subsection{Recirculation versus temperature tuning}
\label{app_sec:tempsweep}

\begin{figure}[t!]
    \centering
    \includegraphics[width=5.5in]{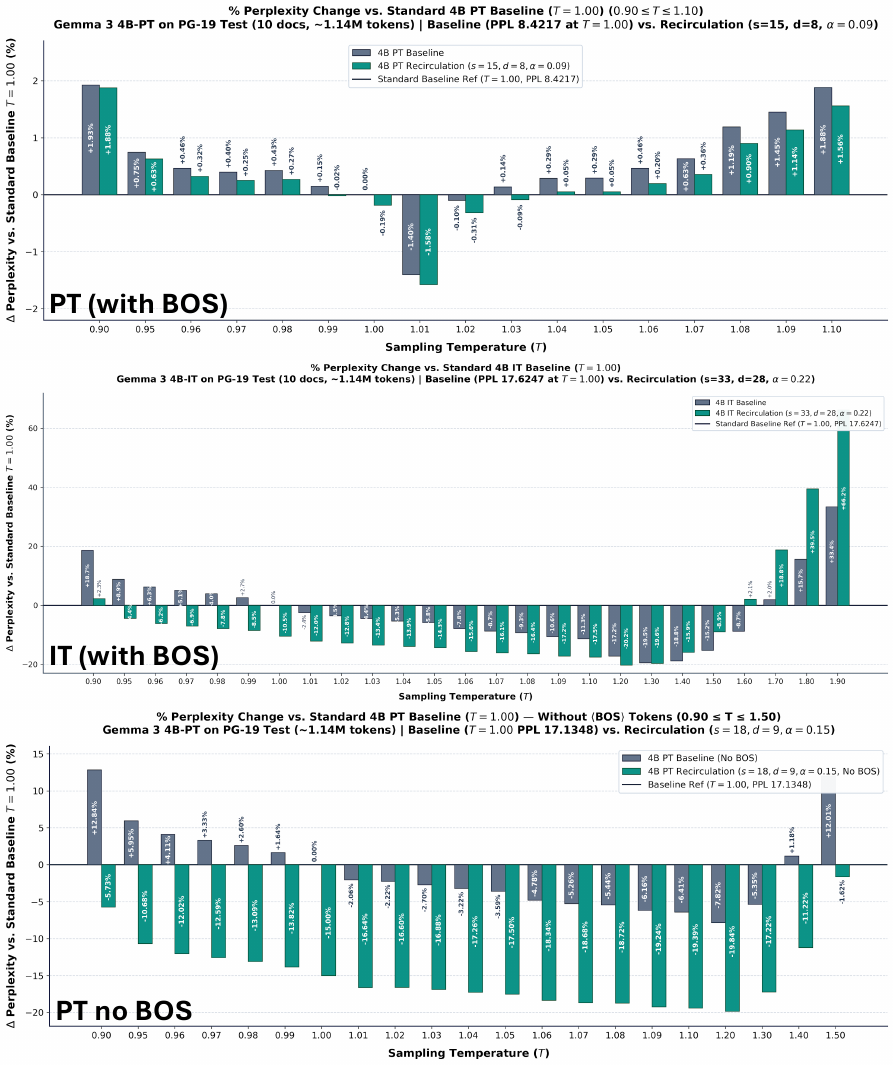}
    \caption{Percentage change in perplexity on PG-19 eval dataset relative to baseline Gemma3 4B model
    for a range of softmax temperatures, both with (green) and without (grey) recirculation.
    (top panel) 4B PT,
    (middle panel) 4B IT,
    (bottom panel) 4B PT where beginning-of-sequence (BOS) tokens are omitted.
    }
    \label{app_fig:tempsweep}
\end{figure}

Gemma3 4B with the PG-19 evaluation (`test') set was used for temperature-tuning experiments. The documents were split into  1024 token chunks per context window and a total of 1.14M tokens were scored.
We swept over a range of softmax temperatures to determine the qualitative effect of temperature 
on perplexity and its interaction with recirculation.
Figure~\ref{app_fig:tempsweep} shows temperature sweeps for the PT model (top panel), the IT model (middle), and
the PT model with BOS tokens removed from the beginning of sequence (bottom). The green bars are with
recirculation; the grey bars are the baseline i.e., the off-the-shelf model.  Lower bars indicate a 
larger percentage reduction in perplexity. For all three variants, recirculation with the 
optimal temperature adjustment (the lowest green bar) reduces perplexity more than the
baseline model with the optimal temperature adjustment (the lowest grey bar), indicating that 
recirculation cannot be discounted as a manipulation of output entropy. (As it turns
out, for all three variants and both baseline and recirculation models, increasing the entropy via $T>1$
is ideal.) For the BOS-absent sequences, recirculation achieves a far greater perplexity reduction than
temperature tuning alone---an additional 12\%---whereas the additional reduction for recirculation
is modest for the BOS-present sequences---0.18\% for the PT and 0.7\% for the IT.

\subsection{Recirculation versus looping}

The dataset is comprised of 250 documents from the arxiv train set, with 2 windows taken from each document starting from the beginning of the document, each of size 1024 tokens. The beginning of the document is marked by a BOS token; the subsequent window is not preceded by a BOS token.
Results for recirculation are based on optimal hyperparameters for the BOS-absent sequences (see
Table~\ref{app_tab:hyperparameters}).

\subsection{Which tokens benefit from recirculation?}
\label{app_sec:which_tokens_benefit}

Experiments in the main paper are based on Gemma3 4B PT with the arXiv data set, training split, with
a context window of 1024 tokens extracted from a randomly selected position within each document with
a BOS inserted at the start of each window. Part-of-speech tags are
extracted with \texttt{nltk.pos\_tag}. We processed 24960 documents
and for each, we recirculated tokens 0-767 individually and examined downstream effects at lags 1-256.

\begin{figure}[tb]
    \centering
    \includegraphics[width=5.5in]{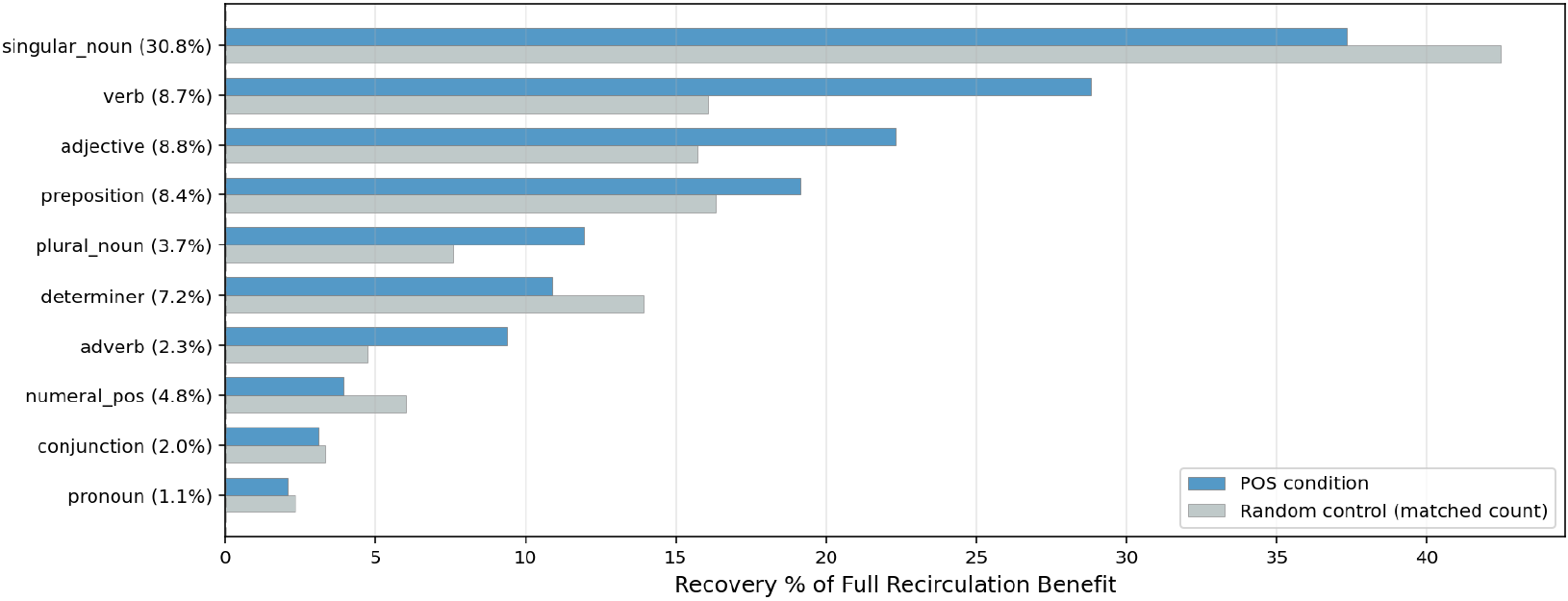}
    \caption{Effect of recirculating all tokens by a specific part-of-speech (blue bars) or count matched random tokens (grey bars).}
    \label{fig:PoS_recirculation}
\end{figure}

We also ran an experiment in which we recirculated all and only tokens tagged with a 
given part of speech (Figure~\ref{fig:PoS_recirculation}).
For this experiment, 3120 documents from the arXiv train set were used,
and from each, a sequence of 1024 tokens was taken from a random
starting position within the document. Each of these sequences was
tested with recirculation of only PoS-selected tokens.

\section{Downstream evaluations}
\setcounter{table}{0} 
\setcounter{figure}{0} 

\subsection{Instruction following}
\label{app_sec:instruction_following}
We generated 800 queries of the following form:
\begin{lstlisting}
<start_of_turn>user
Let's play a game. I will say two words.
If the first word is a fruit, you say first.
If the second word is a fruit, you say second.
For example, if I say `pomegranate kangaroo', respond first.
And if I say `koala fig', respond second.
The word pair is `monkey banana'. Your answer?<end_of_turn>
<start_of_turn>model
\end{lstlisting}
On half the trials, the model was asked instead to identify the position of the \emph{animal} (``If
the first word is an animal, you say first...''). We formed 400 trials by combining twenty different
fruit names with twenty different animal names. With the respond-to-fruit and respond-to-animal variants,
this yields 800 trials total.
The fruits are:
apple, avocado, banana, blueberry, cantaloupe, cherry, grape, honeydew, kiwi, lemon, lime, mango, orange, peach, pear, pineapple, plum, raspberry, strawberry, watermelon.
The animals are:
bear, bird, cat, deer, dog, dolphin, elephant, fox, giraffe, lion, lizard, monkey, penguin, shark, snake, spider, tiger, whale, wolf, zebra.

We selected the most likely response among eight candidate tokens, which consisted of the words
first and second, both in upper- and lower-case form and with and without a leading space.

\begin{figure}[tb]
    \centering
    \includegraphics[width=5.5in]{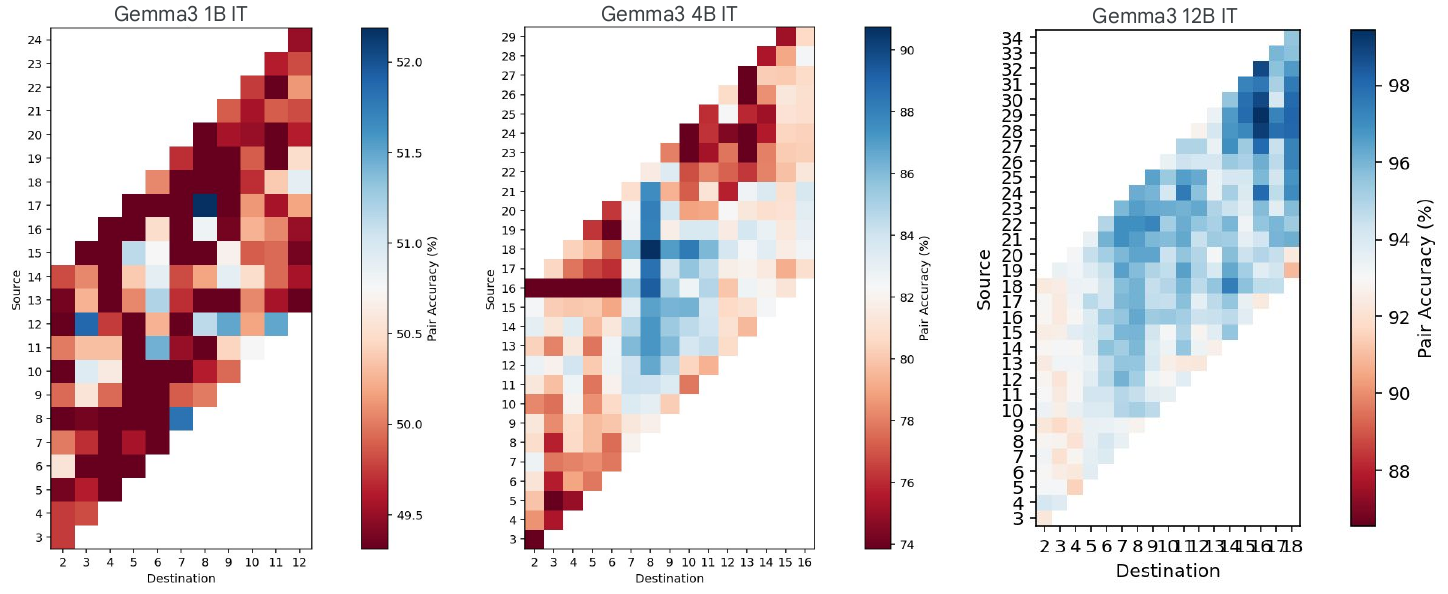}
    \caption{Hyperparameter sweeps for Gemma3 1B, 4B, 12B IT models on the instruction-following task}
    \label{app_fig:instruc_sweep}
\end{figure}

For the Gemma3 4B and 12B IT, we used the source and destination layers determined by our previous
perplexity hyperparameter sweep, and $\alpha=0.07$. For the task-specific result, we conducted
a sweep using the instruction-following dataset to determine an upper bound on performance. These
sweeps are shown in Figure~\ref{app_fig:instruc_sweep}. The 1B sweep is included as 
well, revealing  that the model is essentially performing at chance. For the 4B model, 
source layer 18, destination layer 8 was best; for the 12B model, source layer 29, 
destination layer 16 was best.

\subsection{Contextualization}
\label{app_sec:contextualization}
For Figure~\ref{fig:racing_thoughts} of the main paper, we used hyperparameters chosen based
on minimizing perplexity of the pretrained model. Because performance on the 12B model is near ceiling,
Figure~\ref{fig:racing_thoughts} fails to convey the significant reductions in error, which we highlight in Table~\ref{app_tab:recirculation_results_12b}.
\begin{table}[tb]
\centering
\small
\begin{tabular}{lccr}
\toprule
\textbf{Dataset /} & \textbf{Baseline} & \textbf{Recirculation} & \textbf{Relative} \\
\textbf{\# Distractor Sent.} & \textbf{Error (\%)} & \textbf{Error (\%)} & \textbf{Reduction (\%)} \\
\midrule
\textit{Factual} & & & \\
0    & 0.00  & 0.00  & N/A \\
1    & 0.60  & 0.47  & 22.22\% \\
2    & 0.63  & 0.49  & 21.43\% \\
3    & 0.54  & 0.30  & 43.75\% \\
4    & 0.38  & 0.24  & 35.71\% \\
5    & 0.63  & 0.38  & 39.28\% \\
\midrule
\textit{Gender} & & & \\
0     & 1.25  & 1.25  & 0.00\% \\
1     & 3.75  & 3.33  & 11.11\% \\
2     & 3.06  & 2.50  & 18.18\% \\
3     & 3.23  & 2.40  & 25.81\% \\
4     & 3.58  & 2.58  & 27.91\% \\
5     & 1.81  & 1.18  & 34.62\% \\
\midrule
\textit{Polysemous} & & & \\
0 & 12.50 & 12.50 & 0.00\% \\
1 & 20.42 & 19.17 & 6.12\% \\
2 & 27.78 & 27.22 & 2.00\% \\
3 & 31.46 & 30.21 & 3.97\% \\
4 & 34.17 & 31.33 & 8.29\% \\
5 & 38.47 & 36.25 & 5.78\% \\
\bottomrule
\end{tabular}
\caption{Comparison of Baseline vs. Recirculation Error Rate Reduction across Factual, Gender, and Polysemous datasets for the Gemma3 12B model.}
\label{app_tab:recirculation_results_12b}
\end{table}

\subsection{Multiple-choice and single-token response tasks}
\label{app_sec:mc}
\begin{table*}[htbp]
\centering
\tiny
\setlength{\tabcolsep}{0pt}
\begin{tabular*}{\linewidth}{@{\extracolsep{\fill}} l *{10}{c} @{}}
\toprule
\textbf{Dataset} 
& \multicolumn{2}{c}{\textbf{sou 16 dest 5 $\alpha$ .09}} 
& \multicolumn{2}{c}{\textbf{sou 15 dest 8 $\alpha$ .09}} 
& \multicolumn{2}{c}{\textbf{sou 18 dest 6 $\alpha$ .06}} 
& \multicolumn{2}{c}{\textbf{sou 16 dest 8 $\alpha$ .05}} 
& \multicolumn{2}{c}{\textbf{sou 15 dest 8 $\alpha$ .05}} \\
\cmidrule(lr){2-3} \cmidrule(lr){4-5} \cmidrule(lr){6-7} \cmidrule(lr){8-9} \cmidrule(lr){10-11}
& \textbf{Baseline} & \textbf{Recirculation} 
& \textbf{Baseline} & \textbf{Recirculation} 
& \textbf{Baseline} & \textbf{Recirculation} 
& \textbf{Baseline} & \textbf{Recirculation} 
& \textbf{Baseline} & \textbf{Recirculation} \\
\midrule
MMLU (0-shot)          & 57.90\% & \hlg{58.28\%} & 57.93\% & \hlg{58.12\%} & 57.92\% & \hlg{58.05\%} & 57.92\% & \hlg{58.06\%} & 57.93\% & \hlg{58.25\%} \\
HellaSwag (0-shot)     & 75.89\% & \hlr{75.86\%} & 75.94\% & \hlg{76.17\%} & 76.51\% & \hlg{76.57\%} & 76.51\% & \hlg{76.64\%} & 76.51\% & \hlg{76.68\%} \\
WinoGrande (0-shot)    & 69.46\% & \hlr{69.38\%} & 69.61\% & \hlg{69.85\%} & 69.53\% & \hlg{69.93\%} & 69.53\% & \hlg{69.77\%} & 69.61\% & \hlg{70.48\%} \\
ARC Easy (0-shot)      & 81.78\% & \hlg{82.07\%} & 81.52\% & \hlg{82.11\%} & 82.37\% & \hlg{82.45\%} & 82.37\% & \hlr{82.28\%} & 82.32\% & \hlg{82.62\%} \\
ARC Challenge (0-shot) & 54.44\% & \hlg{54.86\%} & 54.35\% & \hlg{54.44\%} & 54.61\% & \hlg{54.86\%} & 54.61\% & \hlr{54.52\%} & 54.61\% & \hlg{54.95\%} \\
PIQA (0-shot)          & 79.98\% & \hlg{80.52\%} & 79.87\% & \hlg{80.03\%} & 80.03\% & \hlg{80.25\%} & 80.03\% & \hlg{80.14\%} & 80.03\% & \hly{80.03\%} \\
BoolQ (0-shot)         & 79.08\% & \hlg{79.11\%} & 79.08\% & \hlr{78.99\%} & 78.87\% & \hlg{79.54\%} & 78.87\% & \hlg{78.99\%} & 78.84\% & \hlg{79.27\%} \\
Lambada (0-shot acc)   & 70.02\% & \hlg{70.27\%} & 70.06\% & \hlr{69.80\%} & 69.98\% & \hlr{69.92\%} & 69.98\% & \hlg{70.43\%} & 69.96\% & \hlg{70.15\%} \\
\bottomrule
\end{tabular*}
\caption{Zero-shot performance comparison between Baseline and Recirculation across five model configurations. 
The model configurations were chosen based on the hyperparameter sweep to all be good candidates.
We report separate baselines for each run because they were performed on different hardware, which affects detailed model performance.}

\label{app_tab:mc_indiv_runs}
\end{table*}

\begin{figure}[tb]
    \centering
    \includegraphics[width=5.5in]{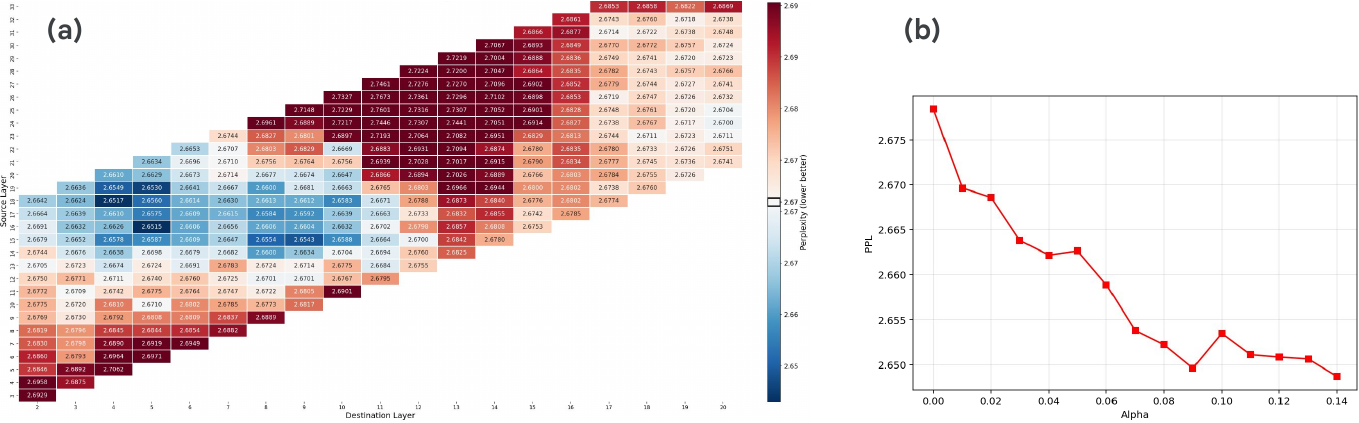}
    \caption{(a) Perplexity of the response token for the 1531 problems
    in the MMLU development set for the Gemma3 4B PT model with recirculation. A sweep is conducted over source and destination layers, fixing 
    $\alpha=.07$;  (b)
    Sweep of $\alpha$, fixing source and destination at 16 and 5, the condition that yields the best performance in (a).}
    \label{app_fig:MMLU_hyperparameter_sweep}
\end{figure}

We tested the Gemma3 4B PT model. To determine the optimal hyperparameters, we
conducted a (source, destination) sweep using the 1531 MMLU development-set
problems. These problems are distinct from the examples used for evaluation.
Figure~\ref{app_fig:MMLU_hyperparameter_sweep}a shows the sweep, fixing
$\alpha=0.07$.  Figure~\ref{app_fig:MMLU_hyperparameter_sweep}b shows a scan
over $\alpha$, fixing source and destination layers to be the pair that yields
the lowest perplexity in the sweep of Figure~\ref{app_fig:MMLU_hyperparameter_sweep}a.
The resulting hyperparameters that were used in the various single-token response
datasets were source 16, destination 5, $\alpha=0.09$.

The evaluations with these hyperparameters are shown in Table~\ref{tab:single_token_benchmarks} and in the first pair of columns in Table~\ref{app_tab:mc_indiv_runs}.
To assess the robustness of these hyperparameters, we tested four candidates drawn from our original perplexity sweep (Figure~\ref{fig:hyperparm_sweep}).
While the remaining hyperparameters did not perform as well as the ones chosen based on the MMLU development set, improvements were reliable on average.
Of the eight datasets and five hyperparameter configurations, only 7 of 40 baseline-recirculation pairs obtain worse performance for recirculation.
A mixed-effects model with crossed random factors (dataset $\times$ configuration) indicates that the gain from
recirculation is highly reliable ($z=5.76$, $p<0.001$ by a two-tailed test).

\subsection{Standard benchmark datasets: GSM8k and HumanEval}
\label{app_sec:gsm8k}

For recirculation, we used 4B model hyperparameters based on sweep with the MMLU development set  (see Table~\ref{app_tab:hyperparameters}). For adaptive recirculation (see Section~\ref{subsec:adaptive_recirc}), we masked the prompt part of the question, and only trained the model to predict the ground-truth response present in the dataset.
Figure~\ref{fig:gsm_results} reports results for adaptive recirculation using our best-performing conditional $\boldsymbol{\alpha}, \boldsymbol{\beta}$ scheme as highlighted in Figure~\ref{fig:adaptive_recirc_ablation}.

As HumanEval is an evaluation benchmark consisting of only 164 examples, we used a larger separate dataset for training, i.e., BigCodeBench \citep{zhuo2025bigcodebench}. BigCodeBench, similar to HumanEval, focuses on code completion given a function signature and docstring.

\subsection{Adaptive recirculation}
\label{app_sec:adaptive_recirculation}

For all training experiments,  we use 250 documents from each of the PG19, C4, and arXiv training sets. 
Documents are partitioned into
windows of 1024 tokens and only completely full windows were included in the training set. 
We use all windows from each document, the first one starts with a BOS token, the remaining ones do not.
Omitting BOS tokens yields higher model perplexity, which facilitates comparison among alternative methods 
for their effectiveness in reducing perplexity. Omitting the BOS tokens thus serves as a helpful diagnostic.
We fix the model hyperparameters based on the BOS-absent Gemma3 1B sweep optimum (Table~\ref{app_tab:hyperparameters}).
The transformer with recirculation and the MLP module is trained with Back Propagation Through Time (BPTT). For the
perplexity studies of Figure~\ref{fig:adaptive_recirc_ablation}, we use the Gemma3 1B PT model with the previously selected
source and destination layers (Table~\ref{app_tab:hyperparameters}).

\begin{figure}[b!]
    \centering
    \includegraphics[height=1.25in]{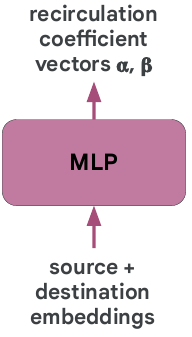}
    \caption{Adaptive recirculation MLP}
    \label{app_fig:adaptive_ppl}
\end{figure}

Our architecture is adapted from the next-latent prediction MLP in \citet{teoh2025}, 
where we use a 2 hidden layer GELU-based MLP with the same hidden size as 
the model dimension, with layer-norm at the input of the MLP
(Figure~\ref{app_fig:adaptive_ppl}). The input is twice the model hidden dimension 
as we concatenate the source and the destination embeddings to be fed to the MLP.
Output size is dependent on the formulation used; e.g., it would be twice the model hidden dimension for \textit{learned conditional vectors} scheme, which predicts two scalars ($\alpha$, $\beta$) per dimension.

We use a logistic activation at the output to ensure that these coefficients lie in [0,1].
Further, we initialize the parameters of the network such that it starts with $\alpha=0.1$ and $\beta=0.9$ at initialization, based on the range of values we found to be suitable from our grid-search results.

We train the model for 100 steps with a batch size of 32 using AdamW \citep{loshchilov2018decoupled}, a learning rate of 3e-4, and a weight decay of 1e-4.
For all simulations other than LLM fine-tuning, we freeze the parameters of Gemma3. For LLM fine-tuning, we disable weight decay and use a small learning rate of 1e-5.
Similarly, for the unconditional prediction schemes (same for all inputs), we disable weight decay and increase the learning rate to 1e-1.

Evaluation was performed with the validation or test set of nine datasets: ArXiv, PubMed, PG19, BookSum, Lambada,
Gov Report, BillSum, OpenWebText, and Big Patent. Note that nonoverlapping subsets of ArXiv and PG19 were used for training
and evaluation.

When training our conditional vector $\boldsymbol{\alpha}, \boldsymbol{\beta}$ scheme on downstream tasks with Gemma3 4B (Table~\ref{tab:single_token_benchmarks} and Figure~\ref{fig:gsm_results}), we use the same hyperparameters as our perplexity experiments.
However, we mask the prompt and only train on the response, which refers to only a single token in the single-token benchmarks (Table~\ref{tab:single_token_benchmarks}).
Because the ARC datasets have fewer examples than the MMLU datasets (MMLU train set: 99842, MMLU test set: 14042, ARC easy: 2251, ARC challenge: 1119), we used multi-epoch training for the ARC datasets as we train on a total of 3200 examples (100 steps with a batch-size of 32).

\end{document}

%% file: recirculation_equations.tex
%

\begin{table}[htbp]
\centering
\renewcommand{\arraystretch}{2.2}
\begin{tabular}{>{\raggedright\arraybackslash}p{6cm} >{\raggedright\arraybackslash}p{9.5cm}}
\toprule
\textbf{Description} & \textbf{Formula} \\
\midrule
no normalization & 
$\begin{aligned}
d' &= (1-\alpha)d + \alpha s \\
   &= \left(1 - \alpha + \frac{\alpha}{\rho} \cos\theta \right) d + \alpha s_{\perp}
\end{aligned}$ \\
\midrule
convex combination, magnitude ratio & 
$\begin{aligned}
d' &= (1-\alpha)d + \alpha \rho s \\
   &= \left(1 - \alpha + \cos\theta\right) d + \alpha \rho s_{\perp}
\end{aligned}$ \\
\midrule
nonconvex combination, magnitude ratio & 
$\begin{aligned}
d' &= d + \alpha \rho s \\
   &= (1 + \alpha \cos\theta) d + \alpha \rho s_{\perp}
\end{aligned}$ \\
\midrule
convex combination, magnitude ratio reweighting, destination renormalized & 
$\begin{aligned}
d' &= (1-\alpha)d + \alpha \rho s\\
   &= \left(1 - \alpha + \cos\theta\right) d + \alpha \rho s_{\perp} \\
d' &\leftarrow \frac{\|d\|}{\|d'\|} d' \\
\end{aligned}$ \\
\midrule
nonconvex combination, recirculate only component of source orthogonal to destination & 
$\begin{aligned}
d' &= d + \alpha \left(s - \frac{s^T d}{d^T d} d\right) \\
   &= d + \alpha s_{\perp}
\end{aligned}$ \\
\midrule
convex combination (when aligned) or nonconvex (when orthogonal), magnitude ratio reweighting & 
$\begin{aligned}
d' &= d + \alpha \rho \left(s - \frac{s^T d}{d^T d} d\right) \\
   &= d + \alpha \rho s_{\perp}
\end{aligned}$ \\
\midrule
nonconvex combination, magnitude ratio reweighting, scale source by novelty & 
$\begin{aligned}
d' &= d + \alpha \frac{d^T d}{s^T d} s \\
   &= (1 + \alpha) d + \frac{\alpha \rho}{\cos\theta} s_{\perp}
\end{aligned}$ \\
\midrule
rotate destination embedding toward source embedding & 
$\begin{aligned}
d' &= \frac{1}{\sin\theta} \left[ \sin((1-\alpha)\theta) d + \sin(\alpha\theta) \rho s \right] \\
   &= \cos(\alpha\theta) d + \frac{\sin(\alpha\theta)}{\sin\theta} \rho s_{\perp}
\end{aligned}$ \\
\bottomrule
\end{tabular}
\caption{A variety of recirculation update rules, expressed both in terms of destination $d$ and source $s$, as well as in terms of $d$ and $s_{\perp}$, 
the orthogonal component of $s$: $s_{\perp} = s - \frac{s^T d}{d^T d} d$.
Other notation: magnitude ratio
$\rho = \frac{\|d\|}{\|s\|}$, and the angle $\theta = \arccos\left(\frac{s^T d}{\|s\| \|d\|}\right)$.}
\label{app_tab:normalization_schemes}
\end{table}
